\documentclass[journal]{IEEEtran}

\usepackage{amsmath,amsfonts}
\usepackage{algorithmic}
\usepackage{array}
\usepackage{textcomp}
\usepackage{stfloats}
\usepackage{url}
\usepackage{verbatim}
\usepackage{graphicx}
\usepackage{orcidlink}
\usepackage{textcomp}

\usepackage{cite}
\usepackage{amsmath,amssymb,amsfonts}
\usepackage{algorithmic}
\usepackage{graphicx}

\usepackage{nicematrix}

\newcommand{\bigtriangleq}{\mathrel{\raisebox{.1ex}{\scalebox{1.3}{$\triangleq$}}}}

\usepackage{textcomp}

\def\Plus{\texttt{+}}
\def\Minus{\texttt{-}}

\usepackage{witharrows}

\usepackage[flushleft]{threeparttable}
\usepackage{tablefootnote}
\usepackage{array}
\newcolumntype{P}[1]{>{\centering\arraybackslash}p{#1}}
\usepackage{multicol}
\usepackage{multirow}
\usepackage{tabulary}
\usepackage{colortbl}

\usepackage{xcolor}

\definecolor{gray}{RGB}{128, 128, 128}
\colorlet{mygray}{gray!15!white}

\definecolor{mygray2}{RGB}{90, 90, 90}

\definecolor{myblue}{RGB}{167, 198, 237}
\colorlet{myblue}{myblue!20!white}

\definecolor{pink}{RGB}{219, 48, 122}
\colorlet{mypink}{pink!15!white}

\definecolor{mygreen}{RGB}{0, 140, 0}
\definecolor{myred}{RGB}{210, 0, 0}

\usepackage{amsmath}
\usepackage{mathrsfs}

\usepackage{pifont}

\usepackage{makecell}
\usepackage{booktabs}
\usepackage{subcaption}

\usepackage{stmaryrd}
\usepackage{mathabx}
\usepackage{float}
\usepackage[switch,columnwise]{lineno}

\usepackage{arydshln}
\usepackage{url,hyperref,microtype}
\hypersetup{
    colorlinks=true,
    citecolor=blue,
    linkcolor=blue,
    filecolor=magenta,      
    urlcolor=black,
    pdftitle={PainFormer},
}

\def\BibTeX{{\rm B\kern-.05em{\sc i\kern-.025em b}\kern-.08em
    T\kern-.1667em\lower.7ex\hbox{E}\kern-.125emX}}
\usepackage{balance}

\begin{document}
\title{PainFormer: a Vision Foundation Model for Automatic Pain Assessment}

\author{Stefanos Gkikas\,\orcidlink{0000-0002-4123-1302}, Raul Fernandez Rojas\,\orcidlink{0000-0002-8393-4241}, and Manolis Tsiknakis\,\orcidlink{0000-0001-8454-1450}

\thanks{Stefanos Gkikas and Manolis Tsiknakis are with the Hellenic Mediterranean University, Department of Electrical and Computer Engineering, Heraklion, Crete 714 10, Greece and with the Institute of Computer Science, Foundation for Research \& Technology-Hellas, Heraklion, Crete GR-70013 Greece (email: gkikas@ics.forth.gr; tsiknaki@ics.forth.gr).

Raul Fernandez Rojas is with the University of Canberra, Faculty of Science and Technology, Canberra, ACT 2617, Australia (email: raul.fernandezrojas@canberra.edu.au)
}
}


\maketitle

\begin{abstract}
Pain is a manifold condition that impacts a significant percentage of the population. Accurate and reliable pain evaluation for the people suffering is crucial to developing effective and advanced pain management protocols. Automatic pain assessment systems provide continuous monitoring and support decision-making processes, ultimately aiming to alleviate distress and prevent functionality decline.
This study introduces \textit{PainFormer}, a vision foundation model based on multi-task learning principles trained simultaneously on \textit{14} tasks/datasets with a total of \textit{10.9} million samples. 
Functioning as an embedding extractor for various input modalities, the foundation model provides feature representations to the \textit{Embedding-Mixer}, a transformer-based module that performs the final pain assessment.
Extensive experiments employing behavioral modalities--including RGB, synthetic thermal, and estimated depth videos--and physiological modalities such as ECG, EMG, GSR, and fNIRS revealed that \textit{PainFormer} effectively extracts high-quality embeddings from diverse input modalities.
The proposed framework is evaluated on two pain datasets, \textit{BioVid} and \textit{AI4Pain}, and directly compared to \textit{75} different methodologies documented in the literature.
Experiments conducted in unimodal and multimodal settings demonstrate state-of-the-art performances across modalities and pave the way toward general-purpose models for automatic pain assessment.  
\textit{\textcolor{blue}{The foundation model's architecture (code) and weights are available at \href{https://github.com/GkikasStefanos/PainFormer}{\textcolor{blue}{https://github.com/GkikasStefanos/PainFormer}}.}}

\end{abstract}



\section{Introduction}
\IEEEPARstart{P}{ain} is a fundamental evolutionary function, signaling potential harm to the organism or indicating the initiation of illness. It is an essential component of the body's defense mechanism to maintain its integrity \cite{santiago_2022}.
In addition, the biopsychosocial model of pain suggests that physical symptoms of pain are the culmination of a dynamic interplay among biological, psychological, and social factors \cite{cohen_vasem_hooten_2021}, which led Williams and Craig \cite{williams_craig_2016} to propose that pain is \textit{\textquotedblleft a distressing experience associated with actual or potential tissue damage with sensory, emotional, cognitive, and social components\textquotedblright}.
Pain can be classified into three types: nociceptive (resulting from tissue damage), neuropathic (originating from nerve damage), or nociplastic (due to a sensitized nervous system). Each type influences diagnostic and treatment choices at various stages \cite{cohen_vasem_hooten_2021}.
From a time-duration perspective, the two main categories are acute and chronic, the former persisting or repeating for over three months \cite{treede_rief_barke_2015}. 
Acute pain typically results from visible physiological damage due to injury, surgery, illness, trauma, or painful medical procedures. It generally resolves when the underlying cause is treated or healed; however, if unresolved, it can evolve into a chronic condition beyond the initial acute phase. Postoperative pain, which is a type of acute pain, occurs specifically after surgical interventions and represents a significant concern for both patients and healthcare providers, highlighting the importance of effective pain management strategies to aid recovery and prevent the transition to chronic pain \cite{sinatra_2010}. Chronic pain manifests in various forms related to the temporal dimension, such as chronic-recurrent (\textit{e.g.}, migraine headaches) or chronic-continuous (\textit{e.g.}, low back pain) \cite{ruddere_tait_2018}.

Pain is a pervasive medical issue worldwide, impacting up to $30\%$ of the general adult population \cite{cohen_vasem_hooten_2021} and between $83\%$ and $93\%$ of senior adults living in residential care facilities \cite{abdulla_adams_2013}.
According to the Global Burden of Disease (GBD) study, pain is the leading cause of years lived with disability (YLD) \cite{james_abate_2018}, with three of the top contributors being chronic pain conditions: back pain, musculoskeletal disorders, and neck pain \cite{usa_bdc_2013}.
The influence of pain expands beyond individuals to affect society, presenting clinical, economic, and social challenges.
The estimated economic and healthcare costs associated with pain due to reduced work productivity in the United States range from $\$560$ to $\$635$ billion, exceeding the costs of heart disease, cancer, and diabetes combined \cite{gaskin_richard_2012}. Europe shows similar patterns, with direct healthcare expenses and indirect socioeconomic costs of chronic pain accounting for  $3\%$ to $10\%$ of the gross domestic product \cite{breivik_eisenberg_2013}. 
Additionally, in Australia, the average annual cost for one of the $15.4\%$ of residents living with chronic pain is between AU$\$22,588$ and AU$\$ 42,979$ when including non-financial costs \cite{deloitte_australia}.
Pain extends beyond direct impacts on a patient's life, leading to a range of adverse effects, including opioid use, drug overuse, addiction, deteriorating social relationships, and psychological disorders \cite{dinakar_stillman_2016}.
Over the past two decades, the use of prescription opioids has significantly increased in the United States, where the rate of overdose deaths has more than quadrupled from $1999$ to $2016$ \cite{seth_rudd_2018}. Additionally, the side effects of these opioids, including lethargy, depression, anxiety, and nausea, significantly affect workforce productivity and overall quality of life \cite{benyamin_trescot_2008}.

Accurate pain assessment is vital for early diagnosis, disease progression monitoring, and treatment effectiveness evaluation, especially in managing chronic pain \cite{gkikas_tsiknakis_slr_2023}. This importance has led to pain being designated as \textit{\textquotedblleft the fifth vital sign\textquotedblright} in nursing literature \cite{joel_lucille_1999}. 
Furthermore, pain evaluation is essential in physiotherapy, where the therapist externally induces stimuli, and understanding the patient's pain level is needed \cite{badura_2021}.
Objectively evaluating pain is crucial for delivering appropriate care, particularly for vulnerable groups who cannot directly or reliably communicate their pain condition, such as infants, young children, those with mental health conditions, and elderly individuals. 
Various methodologies are employed to assess pain, such as self-reporting, where individuals describe their pain experiences, which is currently regarded as the gold standard \cite{dvader_Bostick_2021}.
Pain evaluation methods in clinical settings offer quantifiable indicators of pain, ranging from the Numeric Pain Rating Scale (NPRS) and Visual Analogue Scale (VAS) to quantitative sensory testing methods such as the pressure pain detection threshold (PPDT) \cite{straatman_lukacs_2022}.
Additionally, behavioral indicators such as facial expressions (like grimacing, open mouth, or lifted eyebrows), vocalizations (such as crying, moaning, or screaming), and body and head movements are important markers \cite{rojas_brown_2023}. 
Physiological measurements, including electrocardiography (ECG), electromyography (EMG), galvanic skin responses (GSR), and respiration rate, also provide essential understandings of the physiological manifestations of pain \cite{gkikas_tsiknakis_slr_2023, farmani_bargshady_2025}. 
Furthermore, brain monitoring techniques such as near-infrared spectroscopy (fNIRS) have shown the capability to detect hemodynamic activity changes related to pain stimuli \cite{rojas_liao_2019, bargshady_aziz_2025}.

Over the past two decades, computational science researchers have dedicated their efforts to developing models and algorithms for advancing automatic pain recognition systems \cite{gkikas_phd_thesis_2025}. The aim is to accurately identify the presence and intensity of pain by analyzing physiological and behavioral responses. The emergence of deep learning and artificial intelligence (AI) methods has further explored these automatic approaches, aiming at interpreting the complex, multifaceted nature of pain \cite{gkikas_tsiknakis_slr_2023}.
Multiple studies highlight the potential of automated systems utilizing behavioral or physiological pain assessment modalities \cite{werner__martinez_2019}. Sario \textit{et al}. \cite{sario_haider_2023} demonstrate the viability of accurately recognizing pain via facial expressions, illustrating their value in clinical settings. Multimodal sensing appears especially promising, exhibiting increased accuracy in pain monitoring systems \cite{rojas_brown_2023}. Additionally, incorporating the temporal dimension of the modalities has been demonstrated to enhance the effectiveness of pain assessment \cite{gkikas_tsiknakis_slr_2023}.
In recent years, affective computing research has increasingly adopted thermal imaging techniques \cite{qudah_2021}, motivated by literature indicating that stress and cognitive load have notable effects on skin temperature \cite{ioannou_2014}. These effects arise from the autonomic nervous system's (ANS) control over physiological signals such as heart rate, respiration rate, blood perfusion, and body temperature, all of which reflect human emotions and affects \cite{qudah_2021}. Moreover, muscle contractions can affect facial temperature by transferring heat to the facial skin \cite{jarlier_2011}. Thus, thermal imaging has emerged as a promising technique for capturing transient facial temperature \cite{merla_2004}. 
However, only a few studies have explored thermal imaging in pain research. In \cite{erel_ozkan_2017}, researchers reported that facial temperature increases after exposure to a painful stimulus, suggesting that thermal cameras might be effective tools for monitoring pain. 
In \cite{haque_2018}, the authors demonstrated that thermal videos achieved similar accuracy to RGB videos in recognizing pain in an automatic pain assessment environment. 
Moreover, our previous study \cite{gkikas_tsiknakis_thermal_2024} introduced synthetic thermal videos through a deep-learning generative process and evaluated their effectiveness in recognizing pain.  
The findings showcased that the performances of the synthetic modality are equivalent to the original RGB videos, while the combination of them holds significant potential.
Furthermore, depth cameras are employed in emotion recognition research because they enable the extraction and use of features related to head pose and 3D facial landmark analysis \cite{savran_gur_2013}. These camera sensors capture body and head dynamics, facilitating the analysis of pixel and depth intensity changes associated with movements and affects \cite{ballihi_lablack_2014, imamura_tashiro_2025}. In addition, depth sensors are more invariant and maintain their effectiveness under poor or uneven lighting conditions \cite{szwoch_pieniazek_2015}. 
The research in \cite{li_dong_lu_2023} demonstrated that depth information combined with RGB enhanced the accuracy of micro-expression analysis and, by extension, emotion recognition tasks, while in \cite{kalliatakis_stergiou_2017}, the authors utilized the 3D positions of facial feature points to identify basic emotions.
In pain research, the adoption of depth modality is limited. The authors in \cite{werner_hamadi_2014} utilized depth map videos to estimate head movements, which enhanced pain recognition performance when combined with facial expressions.

With a new emerging paradigm for building AI systems based on foundation models, there has been a shift towards more adaptable and scalable systems capable of generalizing across various tasks and domains. 
A foundation model refers to any model trained on extensive datasets, typically through self-supervision at scale, which can thereafter be adapted---for instance, fine-tuned---to a diverse array of downstream tasks.
Although foundation models rely on conventional deep learning and transfer learning techniques, their large scale leads to the emergence of new capabilities and improved effectiveness across numerous tasks \cite{bommasani_husdon_2021}.
A plethora of examples have appeared recently in the literature.
For instance, SAM \cite{kirillov_mintun_2023} is a foundation model for image segmentation initially trained from scratch on $11$ million images. In subsequent studies \cite{ma_he_2024,wu_ji_2023}, researchers adapted the SAM model for medical imaging by optimizing it for smaller, specialized datasets.
Furthermore, another significant paradigm shift has occurred with the introduction of generalist models \cite{reed_zolna_2022}---a new type of foundation model trained simultaneously on various tasks under a unified learning policy (usually with supervision). This approach particularly benefits computer vision, where embedding representations can differ significantly across tasks and various vision modalities \cite{awais_naseer_2023}.
In the field of automatic pain assessment, there are approaches that employ pre-trained models. However, they follow the traditional method
of pre-training on a particular larger general dataset and fine-tuned for the specific task of pain assessment.
Studies like those detailed in \cite{haque_2018, rodriguez_cucurull_2022} are founded on transfer learning techniques from facial recognition datasets, whereas others, such as \cite{gkikas_tachos_2024,gkikas_tsiknakis_thermal_2024}, employ multi-stage pre-training methods that progressively learn facial and emotional feature representations.

The present study introduces a multi-task learning-based vision foundation model, \textit{PainFormer}, for automatic pain assessment. To the best of our knowledge, this represents the first effort in pain research to develop and implement a foundation system for pain recognition. Our method is motivated by the principles of \cite{reed_zolna_2022}, training simultaneously across various tasks and datasets and founded on the core concept of foundation approaches where representation learning is performed on large-scale corpora and reused to the downstream tasks, in our case, pain assessment.
The proposed framework consists of \textit{PainFormer}, the primary model, \textit{Embedding-Mixer}, and \textit{Video-Encoder}. These modules are described in detail in Section \ref{methodology}.
The key contributions of this research are threefold: (1) the introduction of a foundation model capable of extracting high-quality embeddings regardless of the input modality, (2) the utilization of synthetic thermal videos and estimated depth videos, and (3) an extensive evaluation of diverse behavioral and physiological modalities in unimodal and multimodal settings aiming for effective pain assessment. 

\section{Related work}
Extensive research has been conducted in the field of automatic pain assessment to explore effective methods for evaluating pain-related conditions.
Unimodal methods exploring specific sensory channels, such as vision or contact sensors, have demonstrated high effectiveness while maintaining a degree of simplicity.
In addition, numerous multimodal approaches have been developed, aiming to integrate diverse information streams. These usually combine behavioral and physiological data or various physiological data sources alone. While often enhancing performance, these methods introduce the challenge of managing the complexity associated with data fusion.

Various compelling methods have been introduced, utilizing video modalities, ranging from hand-crafted feature engineering for facial expressions analysis to employing raw videos with sophisticated deep learning architectures.
Werner \textit{et al}. \cite{werner_hamadi_2014} analyzed facial expressions by extracting features based on point distances and head poses, which were estimated using depth information. By integrating these into a unified feature vector and applying a random forest classifier, they achieved a $76.60\%$ accuracy rate in the binary pain classification task.
Patania \textit{et al}. \cite{patania_2022} calculated fiducial points and applied graph neural networks (GNN) to achieve $73.20\%$ accuracy, while the authors in \cite{gkikas_tsiknakis_embc} employed vision transformers on raw videos to reach an accuracy of $77.10\%$.
Other researchers have made efforts to compute biosignal-related information directly from videos, aiming to enhance assessment accuracy while minimizing the reliance on contact sensors.
Yang \textit{et al}. \cite{yang_guan_yu_2021} leveraged raw videos and three-dimensional convolutional neural networks (3D CNNs) to extract facial features and estimate remote photoplethysmography, effectively combining behavioral and physiological cues for enhanced performance. Similarly, the authors in \cite{huang_dong_2022} used 3D CNNs to extract pseudo-heart rate from videos, achieving over a $10\%$ increase in accuracy by integrating facial features with heart rate data compared to using videos in isolation.

Multiple research efforts have been reporting exploring the use of biosignals as standalone modalities, recognizing their reliability and immunity to the limited or exaggerated expressivity issues presented in video-based methods.
Thiam \textit{et al}. \cite{thiam_bellmann_kestler_2019} developed 1D CNNs to analyze and classify raw ECG signals, achieving modest performance, whereas in \cite{werner_hamadi_2014} leveraged domain-specific electrocardiography features—such as the mean of consecutive heartbeat intervals and the square mean root of successive differences—to attain an enhanced performance of $64\%$.
Interestingly, Huang \textit{et al}. \cite{huang_dong_2022} achieved a $65\%$ accuracy in pain detection using pseudo heart rate extracted from videos, whereas in \cite{gkikas_tachos_2024}, authors utilized solely heart rate from ECGs combined with a transformer-based model to achieve an accuracy of $67.04\%$ for the same task.
Employing EMG signals and computing a series of time and frequency domain features, the authors in \cite{patil_2024} achieved modest performances from this challenging modality. In contrast, Kachele \textit{et al}. \cite{kachele_thiam_amirian_werner_2015}, utilizing GSR signals—recognized as the most informative modality for pain assessment—extracted statistical features such as skewness, kurtosis, and the temporal slope of the signal. They achieved an $81.90\%$ accuracy in the binary pain classification by applying a random forest classifier.

Considering the multidimensional nature of pain, integrating various modalities within a multimodal system presents a promising yet challenging approach. Combining different information sources can significantly improve the accuracy and sensitivity of pain assessment. While individual modalities yield satisfactory predictive performance, their effective combination can lead to enhanced outcomes \cite{werner__martinez_2019}, albeit with notable complexities in achieving optimal integration.
Additionally, leveraging cues from multiple channels could be beneficial and crucial, especially in clinical environments where access to a particular modality might be compromised—for instance, when a patient's rotation obstructs facial visibility.
Thiam \textit{et al}. \cite{thiam_bellmann_kestler_2019} exploited raw ECG, EMG, and GSR signals, forming a 2D representation by concatenating them across the temporal axis and employing a 2D CNN for analysis, which yielded an accuracy of $84.40\%$. In contrast, study reported in \cite{kachele_thiam_amirian_2016} achieved an accuracy of $85.70\%$ using the same modalities but focusing on extracting hand-crafted features, thereby underscoring the significance and value of domain-specific engineering.
Zhi \textit{et al}. \cite{zhi_yu_2019} extracted facial descriptors from videos and a plethora of features from ECG, EMG, and GSR, achieving a notable accuracy of $86\%$ in distinguishing no pain from high pain levels. 
In \cite{gkikas_tachos_2024}, the authors enhanced their results by over $5\%$ by integrating raw facial videos with heart rate data, compared to using the video modality individually. 
Similarly, the research in \cite{kachele_werner_2015} utilized facial landmarks and a comprehensive set of $131$ features from ECG and GSR, including time domain, frequency, and entropy-based metrics, to achieve an improvement of more than $6\%$ over the unimodal video approach.

\begin{figure*}[h]
\begin{center}
\includegraphics[scale=0.527]{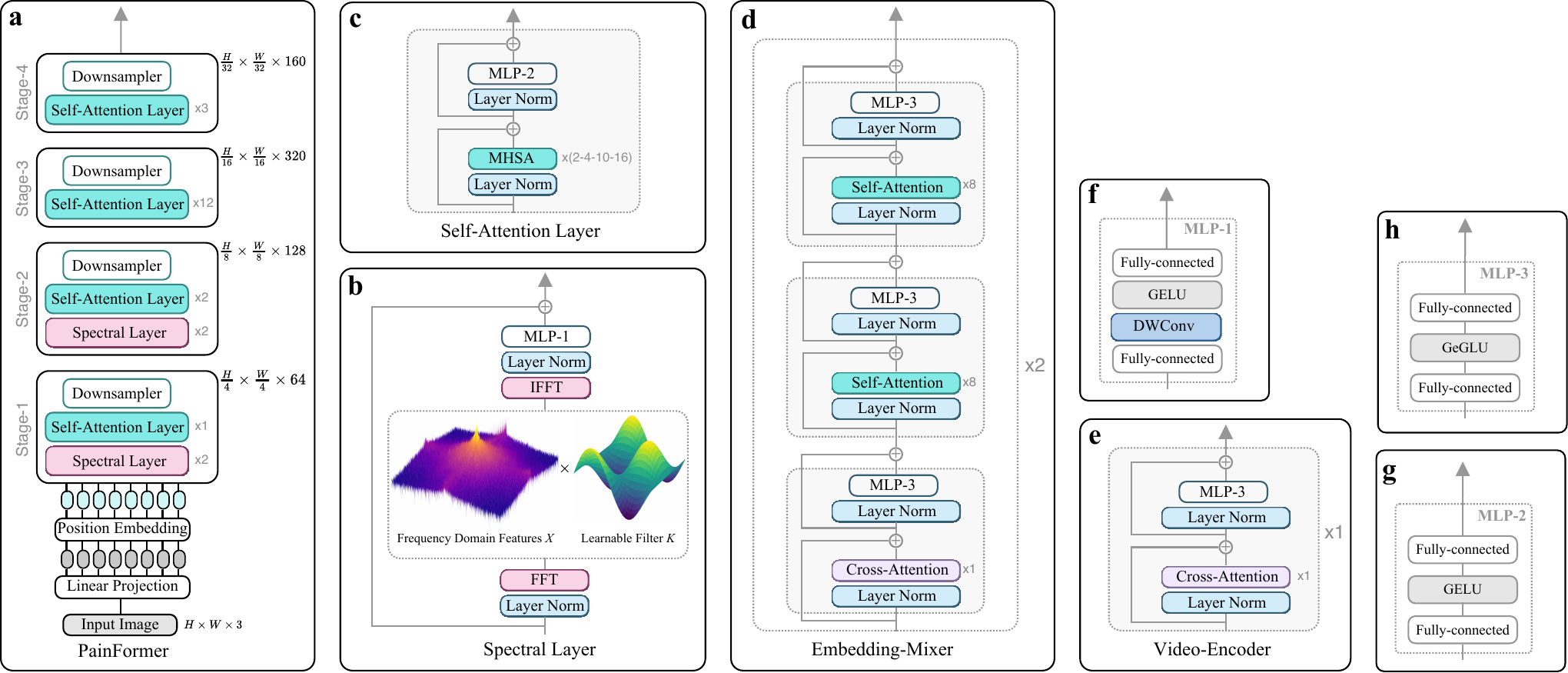} 
\end{center}
\caption{Representation of the main models and their components presented in this study: 
\textbf{(a)} \textit{PainFormer} is organized hierarchically into four stages, integrating \textit{Spectral} and \textit{Self-Attention Layers} to function as the embedding extractor for the inputs; 
\textbf{(b)} The \textit{Spectral Layer}, a primary component of \textit{PainFormer}, applies FFT to compute frequency-related information combined with a learnable filter $K$ to emphasize important frequencies; 
\textbf{(c)} The \textit{Self-Attention Layer}, another primary module of \textit{PainFormer}, facilitates parallel computation of features and their relationship; 
\textbf{(d)} The \textit{Embedding-Mixer}, which combines cross and self-attention mechanisms, serves as the module for final classification of the embeddings used in the pain assessment task; 
\textbf{(e)} The \textit{Video-Encoder}, a compact and efficient module, encodes video representations into a lower dimensional space; 
\textbf{(f)} The \textit{MLP-1} is included in the Spectral Layer; 
\textbf{(g)} The \textit{MLP-2}, part of the \textit{Self-Attention Layer};
\textbf{(h)} The \textit{MLP-3} structure is incorporated within the \textit{Embedding-Mixer} and \textit{Video-Encoder}.}
\label{full}
\end{figure*}

\section{Methodology}
\label{methodology}
This section describes the architecture of the proposed framework's components. It details the foundation model's multi-task learning-based pre-training, augmentation techniques, and training settings for pre-training and pain assessment tasks. Additionally, it explains the generation of synthetic thermal videos, depth estimation for creating depth videos, and visualization of biosignal modalities used in this study.

\subsection{Framework Architecture}
The proposed framework comprises three models: the \textit{PainFormer}, the foundation model that functions as an embedding extractor for inputs; the \textit{Embedding-Mixer}, which utilizes these embeddings, whether individually or combined, for the pain classification task; and the \textit{Video-Encoder}, encoding the video presentations into a lower-dimensional latent space, explicitly employed for the multimodal approach detailed subsequently.
It should be noted that the framework works in two distinct phases rather than in an end-to-end manner: initially, it extracts embeddings, and subsequently, it utilizes them according to the specific modalitie's pipelines and needs.
Table \ref{table:module_parameters} presents the number of parameters and the computational cost of floating-point operations (FLOPS) for the modules and are described in detail in what follows.


\renewcommand{\arraystretch}{1.2}
\begin{table}
\caption{Number of parameters and FLOPS for the modules.}
\label{table:module_parameters}
\begin{center}
\begin{threeparttable}
\begin{tabular}{ P{3.0cm}  P{2.5cm}  P{2.0cm}}
\toprule
Module & Params (Millions) &FLOPS  (Giga) \\
\midrule
\midrule
\textit{PainFormer}      &19.60 &5.82\\
\textit{Embedding-Mixer} &9.85  &2.94 \\
\textit{Video-Encoder}   &3.37  &0.86 \\
\hline
Total &32.82 &9.62\\
\bottomrule
\end{tabular}
\begin{tablenotes}
\scriptsize
\item
\end{tablenotes}
\end{threeparttable}
\end{center}
\end{table}

\subsubsection{PainFormer}
\label{painformer}
Vision Transformers (ViT) have been successfully implemented across various image analysis tasks, showcasing the effectiveness of their core self-attention mechanism \cite{han_xiao_2021}. Furthermore, recent developments in Vision Multilayer Perceptron (Vision-MLP) models that employ spectral mixing techniques—replacing self-attention layers with Fourier transformation layers—demonstrate that simpler architectures with fewer inductive biases can yield comparable outcomes \cite{wave_vision_mlp}.
Our approach incorporates two principal concepts: the hierarchical Vision Transformers (ViT) \cite{swin_transformer}, which utilize multiple stages of embedding extraction to enhance performance and scalability, and the Fourier transform, which, as demonstrated in \cite{guibas_mardani_2021} effectively mixes information from various tokens.
\textit{PainFormer} combines spectral layers, which are implemented using the Fast Fourier Transform (FFT) alongside self-attention layers. 
Specifically, in the early stages of the model, both spectral and self-attention layers are applied. In contrast, the  latter stages exclusively utilize self-attention. 
\hyperref[full]{Fig. 1(a}) illustrates the architecture of the \textit{PainFormer}.

Before discussing the specific components, we provide preliminary information to outline the general concept of the \textit{PainFormer} architecture.
Every 2D input image $I$ is divided into $n$ non-overlapping patches, with each patch $\in\mathbb{R}^{n\times h\times w\times 3 }$, where $h$ and $w$ define the resolution of each patch and are equal to $16 \times 16$, and the $3$ indicates the number of color channels. Each patch is linearly projected into a token with dimension $d = 768$, followed by a positional encoding layer.
Applying Discrete Fourier Transform (DFT), to a 1D input sequence of $N$ elements, $x[n]$, where $n$ ranges from $0$ to $N-1$, results in the following expression:
\begin{equation}
\label{dft}
X[k] = \sum_{n=0}^{N-1} x[n] \, e^{-i 2\pi \frac{k}{N} n}
      = \sum_{n=0}^{N-1} x[n] \, W_N^{kn},
\end{equation}
\noindent
where $i$ denotes the imaginary unit, and $W_N$ denotes the twiddle factor,
$W_N \bigtriangleq e^{-i 2\pi / N}$.
In this manner, the sequence is transformed into the frequency domain. Conversely, the original input sequence can be reconstructed by applying the inverse Discrete Fourier Transform (IDFT):
\begin{equation}
x[n] = \frac{1}{N} \sum_{k=0}^{N-1} X[k] \cdot e^{i2\pi \tfrac{k}{N}n},
\label{idft}
\end{equation}
where $x[n]$ is the initial time-domain sequence.
Furthermore, Eq. (\ref{dft}) can be adapted for two-dimensional inputs, $x[m,n]$, with $0 \leq m \leq M-1$ and 
$0 \leq n \leq N-1$, characterized by:
\begin{equation}
X[u, v] = \sum_{m=0}^{M-1} \sum_{n=0}^{N-1} x[m, n] \cdot e^{-i 2\pi \big(\tfrac{um}{M} + \tfrac{vn}{N}\big)},
\label{2d_dft}
\end{equation}
where $X[u, v]$ represents the two-dimensional frequency representation of the spatial-domain input $x[m,n]$.
\textit{PainFormer} performs the aforementioned processes using specific modules, detailed as follows:

\paragraph{Spectral Layer}
For the tokens $x$ from image $I$, a 2D FFT is applied across the spatial dimensions to transform $x$ into the frequency domain:
\begin{equation}
X = \mathscr{F}[x] \in \mathbb{C}^{h\times w\times d}.
\label{tokens_fft}
\end{equation}
After applying the FFT to extract the various frequency components of the image, we employ a learnable filter, 
$K \in \mathbb{C}^{h\times w\times d}$, which acts as a gate to regulate the significance of each frequency component. This modulation of the spectrum allows for identifying and learning features such as lines and edges. Specifically: 
\begin{equation}
\tilde{X} = K \odot X,
\label{filter_k}
\end{equation}
where $\odot$ defines the element-wise multiplication.
Afterward, the inverse Fast Fourier Transform (IFFT) is applied, which converts the spectral space back into the physical space:
\begin{equation}
x \leftarrow \mathscr{F}^{-1}[\tilde{X}],
\label{ifft}
\end{equation}
where the physical space is referred to as the spatial domain in this case.
The final component of a spectrum layer is an MLP module, which enables efficient channel mixing communication:
\begin{equation}
\Phi(x) = W_2 \cdot \text{GELU}(\text{DWConv}(W_1 \cdot x + b_1)) + b_2,
\label{mlp_spectrum}
\end{equation}
where GELU refers to the Gaussian Error Linear Unit activation function, and DWConv denotes a depthwise convolution layer.
In addition, layer normalization is employed before and after the FFT and IFFT processes, refer to \hyperref[full]{Fig. 1(b)}.

\paragraph{Self-Attention Layer}
The mechanism employed in this layer is the classic self-attention mechanism used in transformers, where for every token sequence $X$, the attention is defined as follows:
\begin{equation}
\text{Att}(X) := \text{softmax} \left( \frac{XW_q (XW_k)^T}{\sqrt{d}} \right) XW_v, 
\label{self_attention}
\end{equation}
where $\text{Att}: \mathbb{R}^{N\times d} \rightarrow \mathbb{R}^{N\times d}$, and $N:= hw$. Also, $W_q$, $W_k$, 
$W_v \in \mathbb{R}^{d\times d}$ are the query, key, and value matrices. 
A layer normalization before and after the attention mechanism similar to the spectrum layer are also applied in the self-attention layer. Furthermore, the MLP module in this layer described as:  
\begin{equation}
\Phi(x) = W_2 \cdot \text{GELU}(W_1 \cdot x + b_1) + b_2.
\label{mlp_attention}
\end{equation}
The architecture of the layer is depicted in \hyperref[full]{Fig. 1(c)}.

\paragraph{Stages} 
A stage-based architecture has been developed to produce a hierarchical representation. \textit{PainFormer} consists of four stages. At the end of each one, a single-layer 2D CNN reduces the number of tokens by downsampling the resolution by a factor of $2$. 
In addition, each stage employs a unique combination of spectral and self-attention layers, varying numbers of heads in the self-attention layers, and different dimensions for the extracted tokens. Table \ref{table:stages} provides the corresponding details.



\renewcommand{\arraystretch}{1.2}
\begin{table}
\caption{Details of the \textit{PainFormer's} architecture.}
\label{table:stages}
\begin{center}
\begin{threeparttable}
\begin{tabular}{ P{0.6cm}  P{1.2cm} P{1.9cm} P{1.9cm} P{1.0cm}}
\toprule
Stage & \# Spectral \newline Layers & \# Self-Attention Layers & \# Self-Attention Heads & Dimension \newline $d$\\
\midrule
\midrule
1  & 2  & 1  & 2  &64 \\
2  & 2  & 2  & 4  &128 \\
3  & -- & 12 & 10 &320 \\
4  & -- & 3  & 16 &160 \\

\bottomrule
\end{tabular}
\begin{tablenotes}
\scriptsize
\item $d$: token dimensions
\end{tablenotes}
\end{threeparttable}
\end{center}
\end{table}

\subsubsection{Embedding-Mixer}
This model is a transformer-based network that incorporates cross- and self-attention mechanisms. As in other studies \cite{jaegle_perceiver_2021}, it introduces an asymmetry in attention computation through cross-attention, which involves fewer latent variables. This approach reduces computational complexity and enhances the model's efficiency. Cross-attention operates similarly to self-attention as in Eq. (\ref{self_attention}). However, the dimensions for 
$W_q$, $W_k$, and $W_v$ are $n\times d$ instead of $d\times d$, where $n<d$ and here, $n$ is set to $256$.
The \textit{Embedding-Mixer} consists of $2$ layers, each containing $1$ cross-attention module and $2$ self-attention modules. 
In addition, the number of heads for the cross and self-attention is $1$ and $8$, correspondingly. The output embedding has a length of $512$ and is utilized for the final classification task, refer to \hyperref[full]{Fig. 1(d)}.

\subsubsection{Video-Encoder}
The architecture of the specific module resembles the \textit{Embedding-Mixer}; however, for efficiency, it comprises just $1$ layer featuring a $1$ cross-attention module with $1$ head. The number of latent variables, $n$, is set to $256$, while the output embedding length is $40$. This module is employed only within the particular framework for one of the multimodal approaches presented, where video embeddings are integrated with GSR embeddings.
The module is illustrated in \hyperref[full]{Fig. 1(e)}.
Note that the \textit{Spectral Layer}, \textit{Self-Attention Layer}, \textit{Embedding-Mixer}, and \textit{Video-Encoder} all incorporate MLP layers, as depicted in \hyperref[full]{Fig. 1(f, g, h)}.


\subsection{Synthetic Thermal \& Depth Videos} 
\label{thermal_depth}
\begin{figure}
\begin{center}
\includegraphics[scale=0.135]{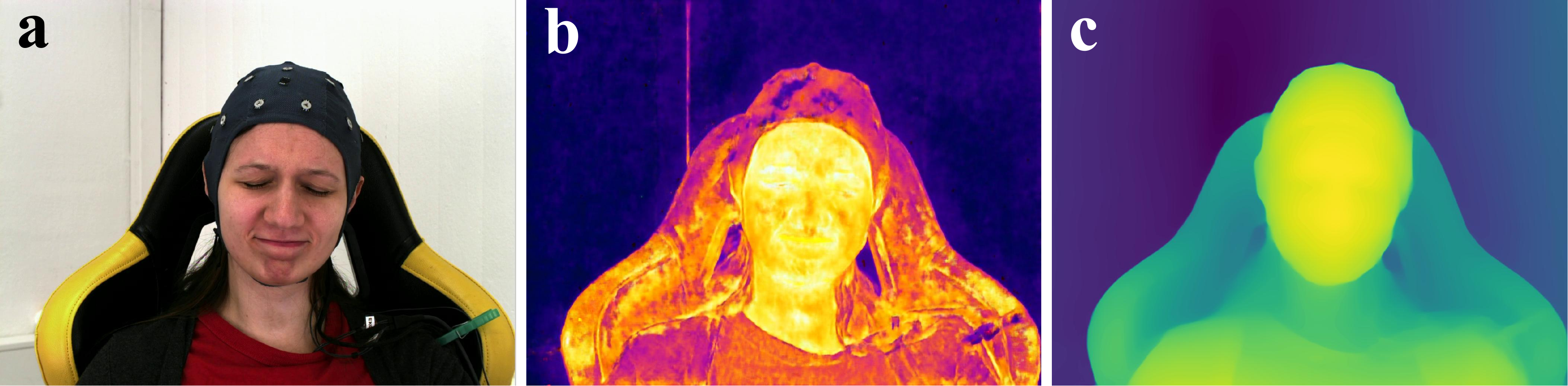} 
\end{center}
\caption{Examples of different vision modalities in frame samples: \textbf{(a)} RGB frame, \textbf{(b)} synthetic thermal frame, and \textbf{(c)} depth estimation frame.}
\label{frame_samples}
\end{figure}

In this study, we integrate thermal and depth vision modalities, in addition to RGB videos, into the pain assessment pipelines. For the thermal modality we employ the thermal videos from our previous work \cite{gkikas_tsiknakis_thermal_2024}, which introduced an image-to-image translation (I2I) approach using a conditional generative adversarial network (cGAN). The network was developed and trained to translate the data distribution from the RGB to the thermal domain, enabling the generation of synthetic thermal representations from new RGB videos. For the depth videos we are using the \textit{\textquotedblleft Depth Anything\textquotedblright} \space method \cite{yang_kang_2024}, a foundational model for monocular depth estimation (MDE) that employs a vision transformer-based encoder-decoder architecture and semi-supervised learning. 
Fig. \ref{frame_samples} presents a frame sample from the RGB, synthetic thermal, and depth modalities.

\subsection{Biosignal Visualization} 
\label{biosignal_visualization}
\begin{figure}
\begin{center}
\includegraphics[scale=0.204]{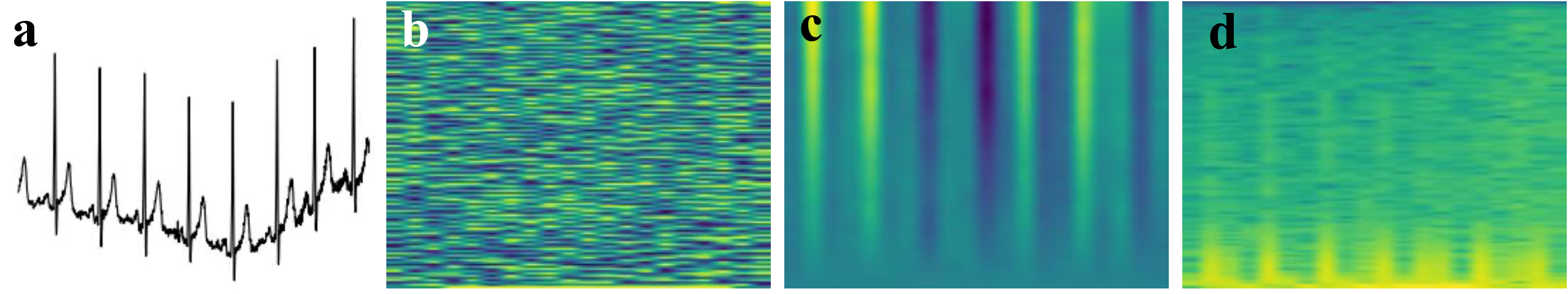} 
\end{center}
\caption{Examples of different visual representations for biosignals: \textbf{(a)} \textit{waveform}, \textbf{(b)} \textit{spectrogram-angle}, \textbf{(c)} \textit{spectrogram-phase}, and \textbf{(d)} \textit{spectrogram-PSD}.}
\label{biosignal_samples}
\end{figure}

Since the foundation model presented in this study is vision-based, a 2D representation of physiological modalities needs to be employed -- representing signals as images has also been successfully demonstrated in other affective-related studies \cite{gkikas_kyprakis_multimodal_2025,gkikas_tsiknakis_painvit_2024, li_zhao_2025}.
Four different visualizations are explored: (1) \textit{waveform} diagrams, which depict the shape and form of a signal as it progresses over time, representing amplitude, frequency, and phase; (2) \textit{spectrogram-angle}, which illustrates the phase angles of the frequencies; (3) \textit{spectrogram-phase}, which shows phase information with unwrapping to address discontinuities; and (4) \textit{spectrogram-PSD}, which represents the power spectral density, exhibiting how power is distributed across frequencies over time. Fig. \ref{biosignal_samples} presents an example of each of these four visualizations.

\subsection{Foundation Training}
\textit{PainFormer}, the proposed foundation model, functions as an embedding extractor, as previously described. To accomplish this, it has been trained extensively across $14$ datasets comprising $10.9$ million samples; refer to Table \ref{table:datasets} for details. 
The training data encompass a range of human-related datasets, including facial recognition datasets like \textit{VGGFace2} \cite{vgg_face_2} and \textit{DigiFace-1M} \cite{digiface1m}  and basic and compound emotion recognition datasets such as \textit{AffectNet} \cite{mollahosseini_hasani_2019} and \textit{RAF-DB} \cite{li_deng_2017}, respectively. Additionally, biosignal datasets, EEG, EMG, and ECG-based, have also been utilized. 
Regarding the training procedure, \textit{PainFormer} has been trained using a multi-task learning approach, where each dataset corresponds to a distinct supervised task. From an architectural perspective, the model remains consistent with the definition in \ref{painformer}. However, it now includes auxiliary classifiers for task-specific purposes. Each classifier is a compact, single-layer, fully connected network with an ELU (Exponential Linear Unit) activation function.
The training objective is to learn from all $14$ datasets/tasks simultaneously. The process is described as follows:
\begin{equation}
L_{total} = \sum_{i=1}^{14} \left[ e^{w_i} L_{S_i} + w_i \right],
\end{equation}
where $L_{S_i}$  represents the loss associated with each specific dataset/task, and $w_i$ denotes the learned weights that enable the learning process by minimizing the aggregate loss $L_{total}$, which encompasses all the individual losses. The foundation model was trained in this manner for $200$ epochs.

\renewcommand{\arraystretch}{1.2}
\begin{table}
\caption{Datasets utilized for the multi-task learning-based pre-training process of the \textit{PainFormer}.}
\label{table:datasets}
\begin{center}
\begin{threeparttable}
\begin{tabular}{ p{3.4cm} p{1.1cm} p{1.0cm} p{1.6cm} }
\toprule
Dataset &\#  samples &\# classes &Modality\\
\midrule
\midrule
\textit{VGGFace2}                              \cite{vgg_face_2}    &3.31M  &9,131   &Facial Images \\
\textit{SpeakingFaces} RGB     \cite{speakingfaces}$^{\varocircle}$ &0.76M  &142     &Facial Images\\
\textit{SpeakingFaces} Thermal \cite{speakingfaces}$^{\varocircle}$ &0.76M  &142     &Facial Images\\
\textit{DigiFace-1M}           \cite{digiface1m}    &0.72M  &10,000  &Facial Images\\
\textit{DigiFace-1M}           \cite{digiface1m}    &0.50M  &100,000 &Facial Images\\

\textit{AffectNet}       \cite{mollahosseini_hasani_2019} &0.40M  &8      &Facial Images\\
\textit{SFace}           \cite{boutros_huber_2022}        &1.84M  &10,341 &Facial Images\\
\textit{CACIA-WebFace}   \cite{yi_lei_2014}               &0.50M  &10,575 &Facial Images\\
\textit{RAF-DB basic}    \cite{li_deng_2017}              &15,000 &7      &Facial Images\\
\textit{RAF-DB compound} \cite{li_deng_2017}              &4,000  &11     &Facial Images\\
\textit{Compound FEE-DB} \cite{du_tao_2014}               &6,000  &26     &Facial Images\\
\textit{EEG-BST-SZ}      \cite{ford_2013}$^{\varodot}$                 &1.50M  &2      &EEG signals\\
\textit{Silent-EMG}      \cite{gaddy_klein_2020}$^{\varodot}$          &0.19M  &8      &EMG signals\\
\textit{ECG HBC Dataset} \cite{kachuee_fazeli_2018}$^{\varodot}$       &0.45M  &5      &ECG signals\\

\Xhline{1.5\arrayrulewidth}
Total: 14 datasets--tasks &10.9M\\
\bottomrule 
\end{tabular}
\begin{tablenotes}
\scriptsize
\item EEG: electroencephalogram \space EMG: electromyography \space ECG \space $\varocircle$: The datasets were also used for the I2I process described in \ref{thermal_depth}, in addition to the training of the \textit{PainFormer} \space $\varodot$: The samples were transformed into spectrograms, equally divided into three parts for each spectrogram type, before being employed.
\end{tablenotes}
\end{threeparttable}
\end{center}
\end{table}

\subsection{Augmentation \& Regularization Methods}
Several augmentation and regularization methods were employed in the \textit{PainFormer's} pre-training and for the downstream task of pain assessment. 
\textit{TrivialAugment} \cite{trivialAugment} and \textit{AugMix} \cite{augmix} were employed for the foundation training. Additionally, a custom augmentation technique involving adjustments to brightness, contrast, saturation, and image cropping was implemented.
The pre-training process also incorporated random noise from a Gaussian distribution. Furthermore, a method was developed to mask out random square sections of input images. \textit{DropPath} \cite{droppath} and \textit{Label Smoothing}
\cite{label_smoothing} were employed to regularize during pre-training.
Two augmentation techniques have been integrated within the framework of the pain assessment task. The first, called \textit{Basic}, combines polarity inversion with noise addition. This approach transforms the original input embeddings by reversing the polarity of data elements and introducing random noise sourced from a Gaussian distribution, thereby creating variability and perturbations. The second technique, \textit{Masking}, applies zero-valued masks to the embeddings, nullifying segments of the vectors. The size and placement of these masks are randomly determined, covering $10\%$ to $20\%$ of the embedding's total dimensions. 
For regularization, the \textit{DropOut} \cite{dropout} and \textit{Label Smoothing} \cite{label_smoothing} techniques were utilized. Table \ref{table:training_details} provides additional details for the two training procedures.

\renewcommand{\arraystretch}{1.2}
\begin{table}
\caption{Training details of the proposed framework.}
\label{table:training_details}
\begin{center}
\begin{threeparttable}
\begin{tabular}{ P{0.50cm} P{0.85cm} P{0.50cm}  P{0.70cm} P{0.75cm} P{0.75cm} P{0.9cm} P{0.5cm}}
\toprule
Task &Optimizer & LR &LR decay &Weight decay &Warmup epochs &Cooldown epochs &Batch size\\
\midrule
\midrule
MTL &\textit{AdamW} &\textit{2e-5} &\textit{cosine}  &0.1 &5 &10 &126$^{\varoast}$ \\
Pain &\textit{AdamW} &\textit{2e-5} &\textit{cosine}  &0.1 &10 &10 &32\\
\bottomrule
\end{tabular}
\begin{tablenotes}
\scriptsize
\item MTL: multi-task learning for pre-training the foundation model \space Pain: pain assessment task \space LR: learning rate \space $\varoast$: batch size is proportionally distributed across the 14 tasks

\end{tablenotes}
\end{threeparttable}
\end{center}
\end{table}

\begin{figure*}
\begin{center}
\includegraphics[scale=0.535]{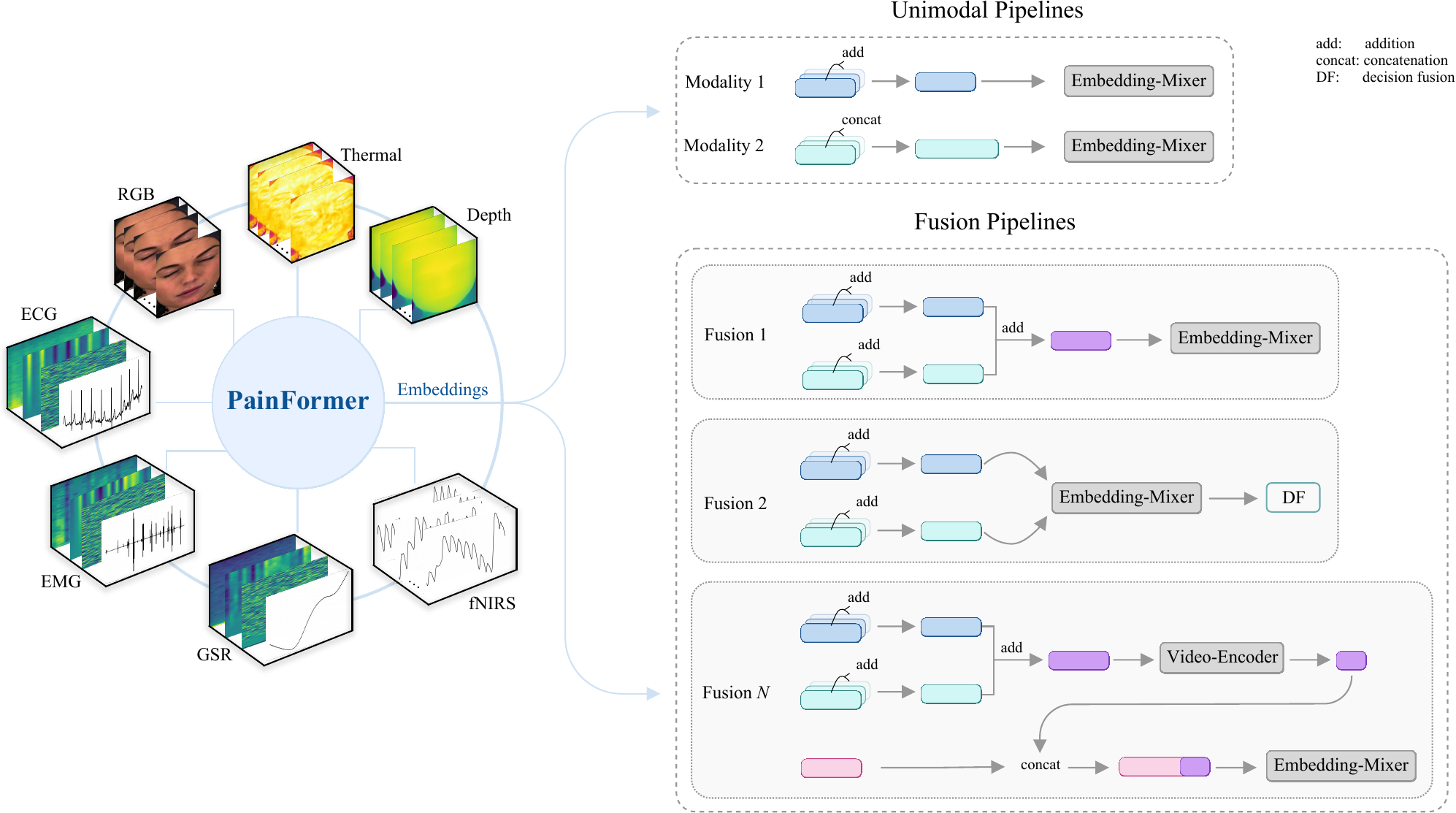} 
\end{center}
\caption{A high-level overview of the presented framework. 
\textit{PainFormer}, the foundation model, is capable of extracting high-quality embeddings from a wide range of different behavioral and physiological modalities. 
Evaluating RGB, thermal, and depth videos and various representations of ECG, EMG, GSR, and fNIRS, including waveforms and spectrograms, demonstrate the comprehensive information encapsulated within these embeddings.
Utilizing the embeddings from the \textit{PainFormer} enables the development of diverse unimodal and multimodal pipelines for the pain assessment task. 
Each pipeline can be customized according to the modalities used, dataset characteristics, and the requirements of the target application or clinical environment.
Our evaluations involved developing and applying various pipelines in unimodal and multimodal settings, achieving state-of-the-art results across different modalities and data representations.}
\label{pipeline}
\end{figure*}

\subsection{Dataset Details}
To assess the performance and robustness of our proposed framework, we conducted experiments on two distinct pain datasets, namely \textit{BioVid} \cite{biovid_2013} and \textit{AI4Pain} \cite{ai4pain}. These datasets provide a diverse and robust basis for assessing the effectiveness of our model in pain assessment.

\subsubsection{BioVid Heat Pain Database}
The particular dataset is widely known and established in the field of pain research.
It includes facial videos, electrocardiograms, electromyograms, and galvanic skin response measurements from eighty-seven $(n\text{=}87)$ healthy individuals ($44$ males and $43$ females, aged between $20$ and $65$). The experimental setup involved using a thermode to induce pain in the participants' right arm. Before data collection, each participant's pain and tolerance thresholds were determined. These thresholds marked the minimum and maximum levels of pain, with two additional intermediate levels, leading to a total of five distinct pain intensities: No Pain (NP), Mild Pain (P\textsubscript{1}), Moderate Pain (P\textsubscript{2}), Severe Pain (P\textsubscript{3}), and Very Severe Pain (P\textsubscript{4}). The temperature settings used for inducing pain varied within the range from P\textsubscript{1} to P\textsubscript{4} but never exceeded $50.5^\circ\text{C}$. Each participant experienced $20$ pain inductions at each of the four predetermined levels of intensity (P\textsubscript{1} to P\textsubscript{4}). Each stimulus lasted $4s$, followed by a recovery period of $8$ to $12s$.
Additionally, $20$ baseline measurements were taken at $32^\circ\text{C}$ (NP), resulting in $100$ stimulations per participant, delivered randomly. The data was then processed to create segments of $5.5s$  starting $1s$ after reaching the target temperature for each stimulation. This processing yielded $8,700$ samples, each lasting $5.5s$, distributed evenly across the five pain intensity classes for each modality, covering all $87$ subjects.
The videos are recorded at a frame rate of $25$ frames per second (FPS), while the biosignal (ECG, EMG, GSR) recordings are sampled at a rate of $51$2 Hz.
It should be noted that data from \textit{Part A} of the dataset were used, where the biosignals had been pre-filtered according to \cite{werner_hamadi_2014}.

\subsubsection{AI4Pain Dataset}
The AI4Pain Grand Challenge 2024 dataset  is a more recent addition to the field and is specifically designed for advanced pain recognition tasks using fNIRS and facial video data. Sixty-five $(n\text{=}65)$ volunteers, including $23$ females, took part in the experiment, with their ages ranging from $17$ to $52$ years.
While the dataset also includes physiological signals such as photoplethysmography (PPG), electrodermal activity (EDA), and respiration (RESP), these additional data are not yet publicly available. The current version of the AI4Pain Challenge dataset is divided into three parts: training ($41$ Volunteers), validation ($12$ Volunteers), and testing ($12$ Volunteers). 
The fNIRS and video recording setup used in this dataset provides comprehensive data on both brain activity and facial movements. The fNIRS data was recorded using an \textit{Artinis} device (Artinis Medical Systems, Gelderland, the Netherlands). This device measures changes in the concentrations of oxygenated haemoglobin (HBO2) and deoxygenated haemoglobin (HHB) (in $\mu$mol/L). The fNIRS system includes $24$ channels covering the prefrontal cortex, with optodes ($10$ sources and $8$ detectors) separated by $30$ mm. Near-infrared light is emitted by sources with wavelengths of $760$ nm and $840$ nm, with a sampling rate of $50$ Hz. The second sensing technology is a video camera (\textit{Logitech StreamCam}) that captures facial video data and movements at a sampling rate of $30$ FPS.
The \textit{AI4Pain} dataset is stratified into three distinct categories of varying levels of pain intensity: \emph{No Pain}, \emph{Low Pain}, and \emph{High Pain}. Specifically, the dataset consists of $65$ instances (each lasting $60s$) of \emph{No Pain}, $780$ instances of \emph{Low Pain} (each lasting $10$s), and $780$ instances of \emph{High Pain} (each lasting $10$s). The \emph{No Pain} category includes instances from the baseline period at the start of each experiment, providing fNIRS and facial video data for comparison with pain-induced responses. The \emph{Low Pain} category comprises instances of mild pain based on the pain tolerance test, capturing subtle neurological and behavioural changes in the corresponding fNIRS and video data. Finally, the \emph{High Pain} category consists of instances where subjects experienced significant pain, also based on the pain tolerance test, leading to notable physiological and behavioural responses in the fNIRS and video data. 
It should be noted that the raw fNIRS signals were obtained directly from the dataset providers, with no pre-processing or filtering applied.

\section{Experimental Evaluation \& Results}
In the context of the present study, several scenarios were designed, both unimodal and multimodal, and used to evaluate the effectiveness of the proposed foundation model. The objective is to leverage various behavioral and physiological modalities to evaluate \textit{PainFormer's} ability to extract and deliver high-quality embeddings for pain assessment. The experimental setup incorporates a wide array of modalities, including RGB, synthetic thermal imaging, depth videos, and physiological measurements like ECG, EMG, GSR, and fNIRS, which feature both waveform and spectrogram representations. 
Furthermore, particular pipelines were designed to suit either individual modalities or a combination thereof, depending on their specific integration requirements. 
This customization is central to our proposal since different pipelines may need to be developed based on particular needs, available data, or specific application needs. 
We want to be able to provide rich feature representations regardless of the input modality and perform exceptionally well across every modality and scenario.
In Fig. \ref{pipeline}, a high-level overview of the proposed framework is presented.
It should be noted that all images, whether video frames or visual representations of biosignals are standardized to a resolution of $224\times224$ pixels.


From the \textit{BioVid} dataset, we utilized \textit{Part A} for experiments in a binary setting, specifically NP vs. P\textsubscript{4}. 
Validation was conducted using the leave-one-subject-out (LOSO) cross-validation method. 
For the \textit{AI4Pain} dataset, multilevel classification was performed using the three available pain levels. 
The validation protocol employed is the hold-out method, as provided by the challenge organizers.
For both datasets, the classiﬁcation metrics utilized include accuracy,
recall (sensitivity), and F1 score.
Additionally, it should be noted that all conducted experiments adhere to a deterministic approach, ensuring they are not influenced by random initializations each time they are performed. This methodology guarantees that any observed differences in performance are solely the result of specific optimization parameters, modalities, and other deliberate variations rather than randomness.

\subsection{BioVid}
A plethora of experiments were conducted using the \textit{BioVid} dataset. 
For the behavioral modalities beyond the original RGB, synthetic thermal and depth videos were also created in order to introduce additional visual representations, as described in \ref{thermal_depth}. 
Also, regarding biosignals, four different representations for ECGs, EMGs, and GSRs were evaluated,
as presented in Section \ref{biosignal_visualization}. Additionally, combinations of these representations were tested.
These are described in what follows.

\subsubsection{Video} 
\label{biovid_video}
Regarding the behavioral modalities from \textit{BioVid}, the \textit{PainFormer} extracts an embedding of dimensionality $d=160$ for each input frame. The embeddings from the respective frames of a specific video are subsequently concatenated to form a unified embedding representation of the entire video:
\begin{equation}
\mathcal{V}_D = [d_1 \| d_2 \| \cdots \|d_m], \quad D \in \mathbb{R}^{N_1},
\label{eq:videos_biovid}
\end{equation}
where $m$ denotes the number of frames in a video, and $N_1$ represents the dimensionality of the unified embedding, calculated as $m\times d \rightarrow 138\times 160=22,080$.
The resulting embedding is fed into the \textit{Embedding-Mixer} for the final pain assessment. 
Beginning with $200$ epochs and using only augmentation techniques for the RGB videos, an accuracy of $71.83\%$ was achieved, with recall reaching $74.52\%$. Similarly, the thermal and depth videos recorded accuracies of $69.83\%$ and $69.00\%$, respectively.
Increasing the training to $300$ epochs with more intense augmentations and applying \textit{Label Smoothing} for regularization improved RGB accuracy to $72.50\%$, though recall slightly decreased by $0.46\%$. In contrast, performance metrics for the thermal modality declined, suggesting a higher sensitivity to augmentations and regularizations in this synthetic modality. Meanwhile, the depth modality showed improved results, achieving $70.08\%$ accuracy, $71.27\%$ recall, and $69.63\%$ F1 score, indicating a positive response to the adjusted training parameters.
In the final experimental setup, training was extended to $600$ epochs with lighter augmentations at a $0.7$ probability, coupled with $0.1$ for \textit{Label Smoothing} and $0.5$ for \textit{DropOut}. This configuration yielded the highest results for the RGB videos, achieving $76.29\%$ accuracy and $77.56\%$ recall. Notably, the F1 score saw the most significant improvement, with an increase of over $5\%$, reaching $75.56\%$.
Similar patterns, albeit with smaller gains, were observed for the thermal and depth videos in the final experimental configuration. Accuracy reached $71.55\%$ for thermal and $71.67\%$ for depth videos, with recall rates nearly identical at $72.83\%$ and $72.84\%$, respectively. These results underline a consistent improvement across all visual modalities with the adjusted training parameters and increased training time.  
Table \ref{table:video} summarizes all the aforementioned experiments and results, highlighting that the RGB modality outperforms the others while the thermal and depth modalities exhibit similar performance levels.
Additionally, although thermal and depth videos increase performances, the improvements are modest, suggesting that they may have reached their maximum potential results.

\renewcommand{\arraystretch}{1.2}
\begin{table}
\caption{Classification results utilizing the video modality, NP vs. P\textsubscript{4} task, reported on accuracy, recall and F1 score.}
\label{table:video}
\begin{center}
\begin{threeparttable}
\begin{tabular}{ P{0.4cm} P{0.6cm} P{0.4cm} P{1.2cm} P{0.3cm} P{0.6cm} P{0.4cm} P{0.4cm} P{0.4cm} }
\toprule
\multirow{2}[2]{*}{\shortstack{Input}}
&\multirow{2}[2]{*}{\shortstack{Epochs}}
&\multicolumn{2}{c}{Augmentation} 
&\multicolumn{2}{c}{Regularization} 
&\multicolumn{3}{c}{Metric}\\ 
\cmidrule(lr){3-4}\cmidrule(lr){5-6}\cmidrule(lr){7-9}
& &\textit{Basic} &\textit{Masking}  &\textit{LS} &\textit{DropOut} &Acc &Rec &F1 \\
\midrule
\midrule
\multirow{3}[1]{*}{\rotatebox{90}{RGB}}  
&200  &0.5 &0.5\textbar10-20\textbar &0.0 &0.0 &71.83 &74.52 &70.29 \\
&300  &0.7 &0.7\textbar15-20\textbar &0.1 &0.0 &72.50 &74.06 &70.93 \\
&600  &0.5 &0.5\textbar15-20\textbar &0.1 &0.5 &\textbf{76.29} &\textbf{77.56} &\textbf{75.56} \\
\Xhline{1.5\arrayrulewidth}
\multirow{3}[1]{*}{\rotatebox{90}{Thermal}}  
&200  &0.5 &0.5\textbar10-20\textbar &0.0 &0.0 &69.83 &71.51 &69.17 \\
&300  &0.7 &0.7\textbar15-20\textbar &0.1 &0.0 &68.83 &69.77 &68.41 \\
&600  &0.5 &0.5\textbar15-20\textbar &0.1 &0.5 &71.55 &72.83 &71.12 \\
\Xhline{1.5\arrayrulewidth}
\multirow{3}[1]{*}{\rotatebox{90}{Depth}}  
&200  &0.5 &0.5\textbar10-20\textbar &0.0 &0.0 &69.00 &69.44 &67.94 \\
&300  &0.7 &0.7\textbar15-20\textbar &0.1 &0.0 &70.08 &71.27 &69.63 \\
&600  &0.5 &0.5\textbar15-20\textbar &0.1 &0.5 &71.67 &72.84 &71.26 \\

\bottomrule 
\end{tabular}
\begin{tablenotes}
\scriptsize
\item \textit{LS: Label Smoothing} \space For Augmentation and Regularization, the number denotes the
probability of application, while in \textit{Masking}, the number in \textbar \space \textbar \space indicates the size of the mask applied.
\end{tablenotes}
\end{threeparttable}
\end{center}
\end{table}

\subsubsection{ECG} 
The training configuration applied to the video data was replicated for the ECG signals. Furthermore, as previously mentioned, four visual representations were employed.
Similar to a video frame, each representation corresponds to an image of $224\times224$ pixels, from which an embedding with a dimensionality of $d=160$ is extracted before being supplied to the \textit{Embedding-Mixer}.
Starting again with $200$ epochs, using minimal augmentation, and without applying regularization techniques, the \textit{waveform} representation achieved an accuracy of $69.58\%$, with recall and F1 scores reaching $72.67\%$ and $68.10\%$, respectively. The \textit{spectrogram-angle} showed lower performance across all metrics with an accuracy of $65.58\%$, whereas the \textit{spectrogram-phase} recorded better accuracy, outperforming the previous two by $0.5\%$ and $4.5\%$, respectively. 
The \textit{spectrogram-PSD} delivered the best results, achieving $71.08\%$, $73.13\%$, and $70.19\%$ across the three metrics.
The same trend continued in the $300$-epoch configuration, which improved results across all representations and metrics. In the final experimental setup of $600$ epochs, while increases were observed across the board, the \textit{spectrogram-PSD} demonstrated the most significant improvements, nearly $4\%$, achieving $75.49\%$ accuracy, $77.15\%$ recall, and $74.90\%$ F1 score. This suggests that for ECG signals, integrating amplitude and frequency information provided by the PSD representation is the most valuable and effective for this analysis.
Table \ref{table:ecg} presents the results for the ECG modality.

\renewcommand{\arraystretch}{1.2}
\begin{table}
\caption{Classification results utilizing the ECG modality, NP vs. P\textsubscript{4} task, reported on accuracy, recall and F1 score.}
\label{table:ecg}
\begin{center}
\begin{threeparttable}
\begin{tabular}{ P{0.4cm} P{0.6cm} P{0.4cm} P{1.2cm} P{0.3cm} P{0.6cm} P{0.4cm} P{0.4cm} P{0.4cm} }
\toprule
\multirow{2}[2]{*}{\shortstack{Input}}
&\multirow{2}[2]{*}{\shortstack{Epochs}}
&\multicolumn{2}{c}{Augmentation} 
&\multicolumn{2}{c}{Regularization} 
&\multicolumn{3}{c}{Metric}\\ 
\cmidrule(lr){3-4}\cmidrule(lr){5-6}\cmidrule(lr){7-9}
& &\textit{Basic} &\textit{Masking}  &\textit{LS} &\textit{DropOut} &Acc &Rec &F1 \\
\midrule
\midrule
\multirow{3}[1]{*}{\rotatebox{90}{Wave}}  
&200  &0.5 &0.5\textbar10-20\textbar &0.0 &0.0 &69.58 &72.67 &68.10 \\
&300  &0.7 &0.7\textbar15-20\textbar &0.1 &0.0 &71.08 &72.74 &70.22 \\
&600  &0.5 &0.5\textbar15-20\textbar &0.1 &0.5 &73.36 &74.75 &72.52 \\
\Xhline{1.5\arrayrulewidth}
\multirow{3}[1]{*}{\rotatebox{90}{Angle}}  
&200  &0.5 &0.5\textbar10-20\textbar &0.0 &0.0 &65.58 &66.68 &64.89 \\
&300  &0.7 &0.7\textbar15-20\textbar &0.1 &0.0 &66.33 &68.22 &65.22 \\
&600  &0.5 &0.5\textbar15-20\textbar &0.1 &0.5 &68.25 &71.24 &66.99 \\
\Xhline{1.5\arrayrulewidth}
\multirow{3}[1]{*}{\rotatebox{90}{Phase}}  
&200  &0.5 &0.5\textbar10-20\textbar &0.0 &0.0 &70.08 &71.54 &69.40 \\
&300  &0.7 &0.7\textbar15-20\textbar &0.1 &0.0 &72.33 &73.73 &71.69 \\
&600  &0.5 &0.5\textbar15-20\textbar &0.1 &0.5 &72.70 &74.19 &72.14 \\
\Xhline{1.5\arrayrulewidth}
\multirow{3}[1]{*}{\rotatebox{90}{PSD}}  
&200  &0.5 &0.5\textbar10-20\textbar &0.0 &0.0 &71.08 &73.13 &70.19 \\
&300  &0.7 &0.7\textbar15-20\textbar &0.1 &0.0 &71.50 &73.14 &70.18 \\
&600  &0.5 &0.5\textbar15-20\textbar &0.1 &0.5 &\textbf{75.49} &\textbf{77.15} &\textbf{74.90} \\

\bottomrule 
\end{tabular}
\end{threeparttable}
\end{center}
\end{table}

\subsubsection{EMG} 
Regarding the EMG signals, the initial configuration of $200$ epochs yielded similar accuracy results across the \textit{waveform}, \textit{spectrogram-phase}, and \textit{spectrogram-PSD} representations, with scores of $68.75\%$, $68.33\%$, and $69.25\%$. In contrast, the \textit{spectrogram-angle} representation underperformed, recording an accuracy of $66.42\%$, a trend also observed in the ECG modality.
Interestingly, in the subsequent training configuration with extended epochs and more intensive augmentation and regularization, the \textit{spectrogram-angle} representation exhibited a notable decline in performance across all metrics, unlike the other representations. Specifically, in the $300$-epoch configuration, despite some improvement, the angle representation still lagged behind the initial results, achieving an accuracy of $65.32\%$, with recall and F1 scores of $68.15\%$ and $63.17\%$, respectively. This indicates that the phase spectrum without phase unwrapping is not well-suited for pain assessment tasks using EMG signals.
Conversely, the other visual representations consistently showed performance improvements in each configuration. Notably, the \textit{spectrogram-PSD} reached the highest accuracy at $72.10\%$ and an F1 score of $71.82\%$. The \textit{waveform} representation achieved the highest recall at $73.64\%$. 
In Table \ref{table:emg}, the results for the EMG modality are presented.

\renewcommand{\arraystretch}{1.2}
\begin{table}
\caption{Classification results utilizing the EMG modality, NP vs. P\textsubscript{4} task, reported on accuracy, recall and F1 score.}
\label{table:emg}
\begin{center}
\begin{threeparttable}
\begin{tabular}{ P{0.4cm} P{0.6cm} P{0.4cm} P{1.2cm} P{0.3cm} P{0.6cm} P{0.4cm} P{0.4cm} P{0.4cm} }
\toprule
\multirow{2}[2]{*}{\shortstack{Input}}
&\multirow{2}[2]{*}{\shortstack{Epochs}}
&\multicolumn{2}{c}{Augmentation} 
&\multicolumn{2}{c}{Regularization} 
&\multicolumn{3}{c}{Metric}\\ 
\cmidrule(lr){3-4}\cmidrule(lr){5-6}\cmidrule(lr){7-9}
& &\textit{Basic} &\textit{Masking}  &\textit{LS} &\textit{DropOut} &Acc &Rec &F1 \\
\midrule
\midrule
\multirow{3}[1]{*}{\rotatebox{90}{Wave}}  
&200  &0.5 &0.5\textbar10-20\textbar &0.0 &0.0 &68.75 &70.55 &67.93 \\
&300  &0.7 &0.7\textbar15-20\textbar &0.1 &0.0 &69.83 &72.52 &68.68 \\
&600  &0.5 &0.5\textbar15-20\textbar &0.1 &0.5 &72.07 &\textbf{73.64} &71.48 \\
\Xhline{1.5\arrayrulewidth}
\multirow{3}[1]{*}{\rotatebox{90}{Angle}}  
&200  &0.5 &0.5\textbar10-20\textbar &0.0 &0.0 &66.42 &68.57 &65.26 \\
&300  &0.7 &0.7\textbar15-20\textbar &0.1 &0.0 &63.92 &66.33 &62.67 \\
&600  &0.5 &0.5\textbar15-20\textbar &0.1 &0.5 &65.32 &68.15 &63.77 \\
\Xhline{1.5\arrayrulewidth}
\multirow{3}[1]{*}{\rotatebox{90}{Phase}}  
&200  &0.5 &0.5\textbar10-20\textbar &0.0 &0.0 &68.33 &69.75 &67.68 \\
&300  &0.7 &0.7\textbar15-20\textbar &0.1 &0.0 &68.58 &70.00 &67.97 \\
&600  &0.5 &0.5\textbar15-20\textbar &0.1 &0.5 &69.37 &71.17 &68.66 \\
\Xhline{1.5\arrayrulewidth}
\multirow{3}[1]{*}{\rotatebox{90}{PSD}}  
&200  &0.5 &0.5\textbar10-20\textbar &0.0 &0.0 &69.25 &70.38 &68.84 \\
&300  &0.7 &0.7\textbar15-20\textbar &0.1 &0.0 &69.67 &71.06 &69.12 \\
&600  &0.5 &0.5\textbar15-20\textbar &0.1 &0.5 &\textbf{72.10} &72.82 &\textbf{71.82} \\

\bottomrule 
\end{tabular}
\end{threeparttable}
\end{center}
\end{table}

\subsubsection{GSR} 
\label{gsr}
Regarding GSR, we observe an apparent difference in performance among the four representations. \textit{Waveform}-based representations significantly outperform the others, with an initial accuracy of $87.75\%$ in the $200$-epoch configuration, surpassing other metrics by over $14\%$.
During longer training sessions, there is a slight improvement across representations, suggesting that the GSR modality may have reached its performance plateau.
The \textit{spectrogram-phase} is the most informative among spectrograms, achieving final accuracy, recall, and F1 scores of $76.41\%$, $77.23\%$, and $76.47\%$, respectively.
The \textit{waveform} representation is the most effective, achieving the highest scores of $88.99\%$ for accuracy, $89.55\%$ for recall, and $88.88\%$ for the F1 metric.
This variation in performance can be attributed to the inherent nature of the GSR signal. As shown in Fig. \ref{pipeline}, GSR typically appears as a smooth curve with gradual slopes, reflecting the slow and steady changes in skin conductivity due to variations in sweat gland activity triggered by stress or arousal.
In contrast, EMG signals display sharp spikes and erratic fluctuations, indicative of the rapid electrical activity associated with skeletal muscle contractions. Similarly, ECG signals are characterized by distinct cyclical patterns such as the P and T waves and the QRS complex.
This suggests that the absence of more complicated patterns in GSR is not well-suited for spectral and frequency domain analysis, which is better leveraged by spectrograms. However, \textit{waveform} representations yield the best results for GSR and outperform all other modalities and visual representations, underscoring their effectiveness in capturing essential physiological information from GSR signals.
Table \ref{table:gsr} summarizes the results for the GSR modality.

\renewcommand{\arraystretch}{1.2}
\begin{table}
\caption{Classification results utilizing the GSR modality, NP vs. P\textsubscript{4} task, reported on accuracy, recall and F1 score.}
\label{table:gsr}
\begin{center}
\begin{threeparttable}
\begin{tabular}{ P{0.4cm} P{0.6cm} P{0.4cm} P{1.2cm} P{0.3cm} P{0.6cm} P{0.4cm} P{0.4cm} P{0.4cm} }
\toprule
\multirow{2}[2]{*}{\shortstack{Input}}
&\multirow{2}[2]{*}{\shortstack{Epochs}}
&\multicolumn{2}{c}{Augmentation} 
&\multicolumn{2}{c}{Regularization} 
&\multicolumn{3}{c}{Metric}\\ 
\cmidrule(lr){3-4}\cmidrule(lr){5-6}\cmidrule(lr){7-9}
& &\textit{Basic} &\textit{Masking}  &\textit{LS} &\textit{DropOut} &Acc &Rec &F1 \\
\midrule
\midrule
\multirow{3}[1]{*}{\rotatebox{90}{Wave}}  
&200  &0.5 &0.5\textbar10-20\textbar &0.0 &0.0 &87.75 &88.68 &87.56 \\
&300  &0.7 &0.7\textbar15-20\textbar &0.1 &0.0 &88.50 &89.16 &88.34 \\
&600  &0.5 &0.5\textbar15-20\textbar &0.1 &0.5 &\textbf{88.99} &\textbf{89.55} &\textbf{88.88} \\
\Xhline{1.5\arrayrulewidth}
\multirow{3}[1]{*}{\rotatebox{90}{Angle}}  
&200  &0.5 &0.5\textbar10-20\textbar &0.0 &0.0 &73.67 &75.00 &73.26 \\
&300  &0.7 &0.7\textbar15-20\textbar &0.1 &0.0 &73.08 &74.60 &72.66 \\
&600  &0.5 &0.5\textbar15-20\textbar &0.1 &0.5 &73.24 &75.02 &72.83 \\
\Xhline{1.5\arrayrulewidth}
\multirow{3}[1]{*}{\rotatebox{90}{Phase}}  
&200  &0.5 &0.5\textbar10-20\textbar &0.0 &0.0 &75.17 &76.13 &74.79 \\
&300  &0.7 &0.7\textbar15-20\textbar &0.1 &0.0 &75.92 &76.60 &75.57 \\
&600  &0.5 &0.5\textbar15-20\textbar &0.1 &0.5 &76.41 &77.23 &76.47 \\
\Xhline{1.5\arrayrulewidth}
\multirow{3}[1]{*}{\rotatebox{90}{PSD}}  
&200  &0.5 &0.5\textbar10-20\textbar &0.0 &0.0 &72.83 &73.91 &72.34 \\
&300  &0.7 &0.7\textbar15-20\textbar &0.1 &0.0 &73.08 &73.96 &72.68 \\
&600  &0.5 &0.5\textbar15-20\textbar &0.1 &0.5 &73.96 &74.81 &73.50 \\

\bottomrule 
\end{tabular}
\end{threeparttable}
\end{center}
\end{table}

\subsubsection{Fusion} 
Multiple fusion approaches have been explored to determine whether integrating multiple representations or modalities leads to performance enhancements. In this study, a combination of inputs from the same sensor type, such as RGB and depth-estimation videos or ECG \textit{waveforms} and ECG \textit{spectrogram-PSD}, is categorized as an unimodal fusion approach. Conversely, combinations involving inputs from different sensor types, such as GSR and EMG, are classified as multimodal.
There are three primary fusion methods, including two feature fusion techniques and one decision fusion technique. Feature fusion includes addition, where embeddings from inputs are combined before proceeding to the subsequent module, and concatenation, which concatenates them along the \textit{y-axis}.
Decision fusion involves processing each embedding through the final module, the \textit{Embedding-Mixer}, which compiles predictions from each source and aggregates them to produce the final prediction.
All corresponding experiments follow the $600$-epoch training configuration as previously presented for the individual input representations.
Table \ref{table:fusion} presents the results of the fusion approaches that were conducted.

In exploring video modality fusion, four combinations were assessed: RGB and thermal, RGB and depth, thermal and depth, and the aggregation of all three---RGB, thermal, and depth.
For the RGB and thermal video combination, regardless of the fusion technique employed, performance was inferior to using RGB alone, with the highest accuracy of $75.66\%$ attained via decision fusion. Similarly, the RGB and depth video combination yielded the best results with decision fusion, with a $75.53\%$ accuracy rate, yet this still fell short of the performance achieved with RGB alone.
Interestingly, the fusion of thermal and depth videos marked an improvement over using depth alone, which was more effective when used independently. With decision fusion, this combination achieved a $73.02\%$ accuracy rate, surpassing depth alone by $1.35\%$. Other metrics also observed improvements, with recall and F1 scores reaching $74.46\%$ and $72.59\%$, respectively.
Finally, the three video inputs of RGB thermal and depth were the only combination that led to improved results compared to the RGB in isolation. Even though some improvements were observed with the addition method, the best results came through the decision fusion, with accuracy, recall, and F1 scores of $76.55\%$, $77.91\%$, and $76.11\%$, respectively---marking improvements of $0.26\%$, $0.35\%$, and $0.55\%$. 
It is noted that in all experiments involving video representations, decision fusion consistently outperformed the addition method.

Regarding biosignals, experiments were conducted solely with ECG and EMG, focusing on the two most effective representations: \textit{waveform} and \textit{spectrogram-PSD}. Fusion experiments for the GSR were not performed, as the waveform representations significantly outperform other visualizations, rendering further fusion unnecessary.
For the ECG, all fusion methods yielded inferior results compared to the \textit{spectrogram-PSD}, with the exception of the addition method, which showed an improvement of $0.21\%$ in recall. 
In EMG biosignals, enhanced performance was noted across all approaches. The most effective method was concatenation, which led to increases of $0.74\%$, $0.36\%$, and $0.64\%$ in the three metrics, achieving respective scores of $72.84\%$, $74.00\%$, and $72.46\%$.

In our multimodal scenario, physiological and behavioral modalities were combined, utilizing GSR signals for the physiological aspect and RGB, synthetic thermal, and estimated depth videos for the behavioral aspect. The \textit{waveform} representation of the GSR signal was used to extract the corresponding embedding vector, where
the individual video feature representations for RGB, thermal, and depth, described in \ref{eq:videos_biovid}, were aggregated to create a unified vector of dimension $22,080$. 
This vector was then fed into the \textit{Video-Encoder} and encoded into a lower dimensional space of $40$. 
The representations from the GSR  and the videos were then concatenated to create a fused vector with a dimension of $160+40=200$. 
The entire process can be expressed as follows:
\begin{equation}
\mathcal{M}_h = \mathcal{G}_d \| Enc\big[(\mathcal{V}^{\tiny{\text{RGB}}}_D + \mathcal{V}^{\tiny{\text{Thermal}}}_D + 
\mathcal{V}^{\tiny{\text{Depth}}}_D)\big],  \quad h \in \mathbb{R}^{N_2},
\end{equation}
where $\mathcal{G}$ represents the GSR embedding, $\mathcal{M}$ is the final fused vector and $N_2$ equal to $200$.
Fig. \ref{pipeline} (bottom right) depicts the specific process. This combination achieved the highest results in the study, reaching $89.08\%$, $89.88\%$, and $88.87\%$ for accuracy, recall, and F1, respectively. Notably, this combination marginally outperformed the GSR modality when used as an individual input in terms of accuracy and recall.

\renewcommand{\arraystretch}{1.2}
\begin{table}
\caption{Classification results on fusion settings$^*$, NP vs. P\textsubscript{4} task, reported on accuracy, recall and F1 score.}
\label{table:fusion}
\begin{center}
\begin{threeparttable}
\begin{tabular}{ P{1.0cm} P{2.5cm} P{1.0cm} P{0.5cm} P{0.5cm} P{0.5cm} }
\toprule
\multirow{2}[2]{*}{\shortstack{Modality}}
& \multirow{2}[2]{*}{\shortstack{Input}} 
& \multirow{2}[2]{*}{\shortstack{Fusion}} 
&\multicolumn{3}{c}{Metric}\\ 
\cmidrule(lr){4-6}
& & &Acc &Rec &F1 \\
\midrule
\midrule
\multirow{15}[1]{*}{Video}
& \multirow{3}{*}{\raisebox{-1.5\height}{RGB, Thermal}} & Add & 75.09 \textcolor{myred}{\scriptsize \Minus 1.20} & 76.97 \textcolor{myred}{\scriptsize \Minus 0.59} & 73.98 \textcolor{myred}{\scriptsize \Minus 1.58} \\

& & DF & 75.66 \textcolor{myred}{\scriptsize \Minus 0.63} & 77.23 \textcolor{myred}{\scriptsize \Minus 0.33} & 75.08 \textcolor{myred}{\scriptsize \Minus 0.48} \\
\cdashline{2-6}[.5pt/1pt]

& \multirow{3}{*}{\raisebox{-1.5\height}{RGB, Depth}}          &Add &74.93 \textcolor{myred}{{\scriptsize \Minus 1.36}} &76.41 \textcolor{myred}{{\scriptsize \Minus 1.15}} &73.38 \textcolor{myred}{{\scriptsize \Minus 2.18}}\\

&  &DF &75.53 \textcolor{myred}{{\scriptsize \Minus 0.76}} &77.18 \textcolor{myred}{{\scriptsize \Minus 0.38}} &75.00 \textcolor{myred}{{\scriptsize \Minus 0.56}}\\
\cdashline{2-6}[.5pt/1pt]

& \multirow{3}{*}{\raisebox{-1.5\height}{Thermal, Depth}}      &Add &71.44 \textcolor{myred}{{\scriptsize \Minus 0.23}} &73.15 \textcolor{mygreen}{{\scriptsize \Plus 0.31}} &70.73 \textcolor{myred}{{\scriptsize \Minus 0.50}}\\

& &DF &73.02 \textcolor{mygreen}{{\scriptsize \Plus 1.35}} &74.46 \textcolor{mygreen}{{\scriptsize \Plus 1.62}} &72.59 \textcolor{mygreen}{{\scriptsize \Plus 1.33}}\\
\cdashline{2-6}[.5pt/1pt]

& \multirow{3}{*}{\raisebox{-1.5\height}{RGB, Thermal, Depth}} &Add &76.26 \textcolor{myred}{{\scriptsize \Minus 0.03}} &77.70 \textcolor{mygreen}{{\scriptsize \Plus 0.14}} &75.78 \textcolor{mygreen}{{\scriptsize \Plus 0.22}}\\

&  &DF &76.55 \textcolor{mygreen}{{\scriptsize \Plus 0.26}} &77.91 \textcolor{mygreen}{{\scriptsize \Plus 0.35}} &76.11 \textcolor{mygreen}{{\scriptsize \Plus 0.55}}\\
\Xhline{1.5\arrayrulewidth}

\multirow{3}{*}{\raisebox{-1.5\height}{ECG}}

& \multirow{3}{*}{\raisebox{-1.5\height}{Wave, PSD}} &Add &75.43 \textcolor{myred}{{\scriptsize \Minus 0.06}} &77.36 \textcolor{mygreen}{{\scriptsize \Plus 0.21}} &74.75 \textcolor{myred}{{\scriptsize \Minus 0.15}}\\

& &Concat &74.74 \textcolor{myred}{{\scriptsize \Minus 0.75}} &76.77 \textcolor{myred}{{\scriptsize \Minus 0.38}} &74.00 \textcolor{myred}{{\scriptsize \Minus 0.90}}\\

\Xhline{1.5\arrayrulewidth}

\multirow{3}{*}{\raisebox{-1.5\height}{EMG}}

& \multirow{3}{*}{\raisebox{-1.5\height}{Wave, PSD}} &Add &72.79 \textcolor{mygreen}{{\scriptsize \Plus 0.69}} &74.15 \textcolor{mygreen}{{\scriptsize \Plus 0.51}} &72.28 \textcolor{mygreen}{{\scriptsize \Plus 0.46}}\\

&  &Concat &72.84 \textcolor{mygreen}{{\scriptsize \Plus 0.74}} &74.00 \textcolor{mygreen}{{\scriptsize \Plus 0.36}} &72.46 \textcolor{mygreen}{{\scriptsize \Plus 0.64}}\\

\Xhline{1.5\arrayrulewidth}
\multirow{2}[1]{*}{Video, GSR}
&RGB, Thermal, Depth, Wave &Add \& Concat &89.08 \textcolor{mygreen}{{\scriptsize \Plus 0.09}} &89.88 \textcolor{mygreen}{{\scriptsize \Plus 0.33}} &88.87 \textcolor{myred}{{\scriptsize \Minus 0.01}}\\

\bottomrule 
\end{tabular}
\begin{tablenotes}
\scriptsize
\item $\ast$: All experiments follow the augmentation and regularization settings for the 600 epoch configuration outlined in the unimodal experiments. \textcolor{mygreen}{+} and \textcolor{myred}{-} indicate an increase or decrease in performance, respectively, compared to the best unimodal input approach. DF: Decision Fusion \space  Add: Addition \space Concat: Concatenation

\end{tablenotes}
\end{threeparttable}
\end{center}
\end{table}

\subsection{AI4Pain}
In the \textit{AI4Pain} dataset, experiments were conducted using unimodal and multimodal approaches. The original RGB videos were utilized for the behavioral modality, while the physiological modality involved waveforms from the fNIRS's HBO2 channels. We note that out of the $24$ available HBO2 channels, $2$ were excluded due to malfunctions.
Table \ref{table:ai4pain_validation} presents the corresponding results. 

\subsubsection{Video}
Similar to \ref{biovid_video}, an embedding of $d=160$ is extracted for every frame, but here, the extracted embeddings are aggregated into a fused vector:
\begin{equation}
\mathcal{V}_d = [d_1 + d_2 + \cdots +d_m], \quad d \in \mathbb{R}^{N_3},
\label{eq:videos_ai4pain}
\end{equation}
where $m$ denotes the number of frames in a video, and $N_3$ represents the dimensionality of the unified embedding, 
equal to $160$.
After feeding the embedding into the \textit{Embedding-Mixer} and following the same $600$-epoch training configuration as in previous experiments, the setup achieved an accuracy of $49.77\%$, with recall and F1 scores of $50.11\%$ and $49.77\%$ respectively. Increasing the \textit{DropOut} rate to 0.3 improved the accuracy and F1 scores to $51.39\%$ and $51.31\%$. Further elevating the \textit{DropOut} to $0.8$ enhanced the recall to $52.74\%$.

\subsubsection{fNIRS}
Similarly, embeddings were aggregated for the $22$ HBO2 channels, resulting in a feature representation of $\mathcal{O}_d=160$. The 600-epoch training setup initially yielded $43.06\%$, $42.80\%$, and $42.07\%$ for the three metrics. By increasing the \textit{DropOut} to $0.3$, a peak performance of $44.44\%$, $45.55\%$, and $43.74\%$ was achieved.

\subsubsection{Fusion}
For the fusion of video and fNIRS data, an aggregation approach was employed:
\begin{equation}
\mathcal{F}_d = \mathcal{V}_d + \mathcal{O}_d, \quad d \in \mathbb{R}^{N_3},
\label{eq:fusion_ai4pain}
\end{equation}
where $\mathcal{F}_d$ is the combined feature representation.
Beginning with the same $600$-epoch training setup, $50.00\%$, $51.01\%$, and $48.54\%$ results were reached for the three metrics. Increasing the \textit{DropOut} to $0.8$ improved the accuracy and F1 score by $0.23\%$ and $1.7\%$, respectively, with the recall decreased by $0.75\%$. The optimal \textit{DropOut} setting of $0.6$ achieved peak performances of $51.85\%$, $51.57\%$, and $51.35\%$ for the accuracy, recall, and F1 scores.

\renewcommand{\arraystretch}{1.2}
\begin{table}
\caption{Classification results on the validation set of \textit{AI4Pain} dataset, multilevel classification task, reported on accuracy, recall and F1 score.}
\label{table:ai4pain_validation}
\begin{center}
\begin{threeparttable}
\begin{tabular}{ P{0.4cm} P{0.6cm} P{0.4cm} P{1.2cm} P{0.3cm} P{0.6cm} P{0.4cm} P{0.4cm} P{0.4cm} }
\toprule
\multirow{2}[2]{*}{\shortstack{Input}}
&\multirow{2}[2]{*}{\shortstack{Epochs}}
&\multicolumn{2}{c}{Augmentation} 
&\multicolumn{2}{c}{Regularization} 
&\multicolumn{3}{c}{Metric}\\ 
\cmidrule(lr){3-4}\cmidrule(lr){5-6}\cmidrule(lr){7-9}
& &\textit{Basic} &\textit{Masking}  &\textit{LS} &\textit{DropOut} &Acc &Rec &F1 \\
\midrule
\midrule
\multirow{3}[1]{*}{\rotatebox{90}{Video}}  
&600  &0.5 &0.5\textbar15-20\textbar &0.1 &0.5 &49.77 &50.11 &49.77\\
&600  &0.5 &0.5\textbar15-20\textbar &0.1 &0.3 &\textbf{51.39} &51.50 &\textbf{51.31} \\
&600  &0.5 &0.5\textbar15-20\textbar &0.1 &0.8 &48.38 &\textbf{52.74} &46.69\\
\Xhline{1.5\arrayrulewidth}
\multirow{3}[1]{*}{\rotatebox{90}{fNIRS}}  
&600  &0.5 &0.5\textbar15-20\textbar &0.1 &0.5 &43.06 &42.80 &42.07 \\
&600  &0.5 &0.5\textbar15-20\textbar &0.1 &0.3 &\textbf{44.44} &\textbf{45.55} &\textbf{43.74} \\
&600  &0.4 &0.4\textbar15-20\textbar &0.1 &0.1 &43.06 &44.18 &42.44 \\
\Xhline{1.5\arrayrulewidth}
\multirow{3}[1]{*}{\rotatebox{90}{Fusion}}  
&600  &0.5 &0.5\textbar15-20\textbar &0.1 &0.5 &50.00 &51.01 &48.54 \\
&600  &0.1 &0.1\textbar15-20\textbar &0.1 &0.8 &50.23 &50.25 &50.24 \\
&600  &0.4 &0.4\textbar15-20\textbar &0.1 &0.6 &\textbf{51.85} &\textbf{51.87} &\textbf{51.35} \\
\bottomrule 
\end{tabular}
\begin{tablenotes}
\scriptsize
\item Fusion: the Addition method of the modalities applied
\end{tablenotes}
\end{threeparttable}
\end{center}
\end{table}

\section{Comparison with existing methods}
To evaluate \textit{PainFormer}, we benchmark our approach with studies from the literature. Specifically, studies that employ the \textit{BioVid} dataset (\textit{Part A}), utilize all available subjects ($87$), conduct the same task, follow the leave-one-subject-out (LOSO) validation protocol, and report accuracy metrics. Similarly, for the \textit{AI4Pain} dataset, our comparisons were with studies that adhered strictly to the evaluation guidelines outlined in the corresponding challenge.

In \textit{BioVid}, the proposed approach utilizing RGB, thermal, and depth video inputs in video-based studies ranks among the highest performance. Achieving an accuracy of $76.55\%$ surpasses all approaches that utilize hand-crafted features, such as those in references \cite{werner_hamadi_2014,werner_hamadi_walter_2017,patania_2022}. Additionally, it outperforms the majority of deep learning-based methods, such as \cite{zhi_wan_2019,thiam_kestler_schenker_2020,tavakolian_bordallo_liu_2020,gkikas_tsiknakis_embc}.
However, there are exceptions, specifically, the results reported in \cite{gkikas_tachos_2024} at $77.10\%$, in \cite{alhamdoosh_pala_2025} at $77.30\%$, and in \cite{yang_guan_yu_2021} at $78.90\%$. Moreover, the authors in \cite{huang_dong_2022} achieved $77.50\%$ using a 3D CNN approach. Combined with the pseudo heart rate extracted from the videos, they reached the highest reported results reported in the literature at 88.10\%. Table \ref{table:video_based} presents the results. 

Regarding the biosignals, it is noted that in ECG-based studies, \textit{PainFormer} achieved the highest results in the literature, with an accuracy of $75.49\%$ using the \textit{spectrogram-PSD} representation. Compared to the subsequent leading studies  \cite{gkikas_chatzaki_2023,gkikas_tachos_2024}, our approach  shows a significant improvement, outperforming them by more than $6\%$ and $8\%$, respectively.
In the relatively few EMG-based studies employing the addition of \textit{waveform} and \textit{spectrogram-PSD} representations, we yielded a $72.84\%$ accuracy, significantly outperforming the following closest study \cite{werner_hamadi_2014}, which achieved $57.90\%$.
In the GSR-based studies, our approach, utilizing solely \textit{waveform} representation and achieving an $88.99\%$ accuracy, leads to performance. It is also observed that studies using raw biosignals rather than extracting domain-specific features generally exhibit better results. The second \cite{lu_ozek_kamarthi_2023} and third \cite{phan_iyortsuun_2023} ranked studies achieved $85.56\%$ and $84.80\%$ accuracy, respectively.
The Table \ref{table:biosignal_based} presents the corresponding results for the biosignals.

In multimodal approaches, our method utilizing video inputs and GSR achieved an $89.08\%$ accuracy, marking the highest result reported in the literature (refer to Table \ref{table:multimodal_based}). Additionally, with one exception \cite{gkikas_tachos_2024}, all documented studies incorporated the GSR signal as one of the inputs. This is consistent with findings that GSR is the most effective modality for pain assessment. For instance, the second-highest-performing study \cite{jiang_li_he_2024}, which used a combination of GSR and ECG signals, achieved $87.06\%$. At the same time, the authors in \cite{zhi_yu_2019}, which included videos, ECG, EMG, and GSR, reached an accuracy of $86.00\%$.

Finally, concerning the \textit{AI4Pain} dataset, \textit{PainFormer} achieved $53.67\%$ accuracy using the RGB video modality, outperforming \cite{prajod_schiller_2024} with $49.00\%$ but falling short of \cite{nguyen_yang_2024}, which reached $55.00\%$ using a transformer-based masked autoencoder. 
Utilizing only the fNIRS, an accuracy of $52.60\%$ was achieved, which is approximately $1\%$ lower than the $53.66\%$ reported in \cite{khan_aziz_2025}.
In a multimodal approach combining videos with waveform representations, an accuracy of $55.69\%$ was achieved, surpassing \cite{gkikas_kyprakis_multimodal_2025} nearly $1\%$, \cite{gkikas_tsiknakis_painvit_2024} by more than $9\%$, \cite{vianto_2025} by over $4\%$, and \cite{gkikas_kyprakis_eda_2025} (using only EDA) by $0.52\%$, establishing the highest performance reported on this dataset to date.
Table \ref{table:ai4pain_test} presents the corresponding results.

\renewcommand{\arraystretch}{1.2}
\begin{table}
\caption{Comparison of video-based studies utilizing \textit{BioVid (Part-A)}, NP vs. P\textsubscript{4} task and LOSO cross-validation.}
\label{table:video_based}
\begin{center}
\begin{threeparttable}
\begin{tabular}{ P{0.9cm} P{3.5cm} P{2.2cm} P{0.6cm} }
\toprule
\multirow{2}[2]{*}{\shortstack{Study}}
&\multicolumn{2}{c}{Method} 
&\multirow{2}[2]{*}{\shortstack{Acc\%}}\\
\cmidrule(lr){2-3}
&Features &ML &\\

\midrule
\midrule
\cite{zhi_wan_2019}  &raw &SLSTM &61.70\\ \hdashline
\cite{thiam_kestler_schenker_2020} &raw &2D CNN, biLSTM  &69.25\\ \hdashline
\cite{werner_hamadi_walter_2017} &optical flow &RF &70.20\\ \hdashline
\cite{tavakolian_bordallo_liu_2020} &raw &2D CNN &71.00\\ \hdashline
\cite{gkikas_tsiknakis_thermal_2024} &raw$^{\varocircle}$ &Vision-MLP &71.03\\ \hdashline
\cite{huang_xia_li_2019}$^\dagger$  &raw &2D CNN &71.30\\ \hdashline
\cite{werner_hamadi_2014} &facial landmarks,  3D distances &RF &71.60\\ \hdashline
\cite{werner_2016} &facial 3D distances &Deep RF &72.10\\ \hdashline
\cite{werner_2016} &facial action descriptors &Deep RF &72.40\\ \hdashline 
\cite{kachele_werner_2015} &facial landmarks, 3D distances &RF &72.70\\ \hdashline
\cite{patania_2022}  &fiducial points &GNN &73.20\\ \hdashline
\cite{gkikas_tsiknakis_embc} &raw &Transformer &73.28\\ \hdashline
\cite{huang_xia_2020}$^\dagger$  &raw &2D CNN, GRU &73.90\\ \hdashline
\cite{gkikas_tachos_2024} &raw &Transformer &77.10\\ \hdashline
\cite{alhamdoosh_pala_2025} &facial landmarks &STAGCN &77.30\\ \hdashline
\cite{huang_dong_2022}  &raw &3D CNN &77.50\\ \hdashline
\cite{yang_guan_yu_2021} &raw, rPPG\textsuperscript{\ding{70}} &3D CNN &78.90\\ \hdashline
\cite{huang_dong_2022}  &raw,  heart rate\textsuperscript{\ding{72}} &3D CNN &\textbf{88.10}\\ \hdashline
\rowcolor{mygray}Our &raw\textsuperscript{\ding{95}} &Transformer &76.55\\

\bottomrule 
\end{tabular}
\begin{tablenotes}
\scriptsize
\item $\dagger$: reimplemented for pain intensity estimation on \textit{BioVid} by \cite{huang_dong_2022} \space 
$\varocircle$: RGB, synthetic thermal videos \space \ding{70}: remote  photo plethysmography (estimated from videos) \space \ding{72}: pseudo heart rate gain (estimated from videos) \space \ding{95}: RGB-thermal-depth (DF) \space ML: Machine Learning \space SLSTM: Sparse Long Short-Term Memory \space biLSTM: Bidirectional LSTM \space RF: Random Forest \space MLP: Multi-Layer Perceptron \space GNN: Graph Neural Networks \space GRU: Gated Recurrent Unit \space STAGCN: Spatio-temporal Attention Graph Convolution Network

\end{tablenotes}
\end{threeparttable}
\end{center}
\end{table}

\renewcommand{\arraystretch}{1.2}
\begin{table}
\caption{Comparison of biosignal-based studies utilizing \textit{BioVid (Part-A)}, NP vs. P\textsubscript{4} task and LOSO cross-validation.}
\label{table:biosignal_based}
\begin{center}
\begin{threeparttable}
\begin{tabular}{ P{0.7cm} P{0.90cm} P{2.2cm} P{2.2cm} P{0.7cm} }
\toprule
\multirow{2}[2]{*}{\shortstack{Study}}
&\multirow{2}[2]{*}{\shortstack{Modality}}
&\multicolumn{2}{c}{Method} 
&\multirow{2}[2]{*}{\shortstack{Acc\%}}\\
\cmidrule(lr){3-4}
&&Features &ML &\\

\midrule
\midrule

\cite{thiam_bellmann_kestler_2019}  &ECG &raw &1D CNN  &57.04 \\ \hdashline
\cite{patil_2024} &ECG &domain-specific$^\divideontimes$ &LR &57.40\\ \hdashline
\cite{martinez_picard_2018_b}  &ECG &domain-specific$^\divideontimes$ &LR &57.69\\ \hdashline
\cite{pavlidou_tsiknakis_2025} &ECG & domain-specific$^\divideontimes$ &SVM &58.39\\ \hdashline
\cite{gkikas_chatzaki_2022}  &ECG & domain-specific$^\divideontimes$ &SVM &58.62\\ \hdashline
\cite{phan_iyortsuun_2023} &ECG &raw &1D CNN, biLSTM &61.20\\ \hdashline
\cite{werner_hamadi_2014}  &ECG &domain-specific$^\divideontimes$ &RF &62.00\\ \hdashline
\cite{kachele_werner_2015} &ECG & domain-specific$^{\divideontimes}$ &SVM &62.40\\ \hdashline
\cite{patil_patil_2024} &ECG & raw &2D CNN, biLSTM  &63.20\\ \hdashline
\cite{huang_dong_2022}  &ECG &heart rate\textsuperscript{\ding{72}} &3D CNN &65.00\\ \hdashline
\cite{gkikas_tachos_2024} &ECG &heart rate &Transformer &67.04\\ \hdashline
\cite{gkikas_chatzaki_2023}  &ECG &domain-specific$^\divideontimes$ &FCN &69.40\\ \hdashline
\rowcolor{mygray}Our &ECG &raw\textsuperscript{\ding{70}} &Transformer &\textbf{75.49}\\
\Xhline{1.5\arrayrulewidth}

\cite{pavlidou_tsiknakis_2025} &EMG & domain-specific$^\divideontimes$ &LSTM &56.83\\ \hdashline
\cite{werner_hamadi_2014} &EMG &domain-specific$^\divideontimes$ &RF &57.90\\ \hdashline
\cite{patil_2024} &EMG &domain-specific$^\divideontimes$ &LR &58.59\\ \hdashline
\cite{thiam_bellmann_kestler_2019}  &EMG &raw &2D CNN &58.65\\ \hdashline
\rowcolor{mygray}Our &EMG &raw\textsuperscript{\ding{96}} &Transformer &\textbf{72.84}\\
\Xhline{1.5\arrayrulewidth}

\cite{werner_hamadi_2014} &GSR &domain-specific$^\divideontimes$ &RF &73.80\\ \hdashline
\cite{martinez_picard_2018_b} &GSR &domain-specific$^\divideontimes$ &LR &74.21\\ \hdashline
\cite{kachele_werner_2015} &GSR & domain-specific$^\divideontimes$ &RF &74.40\\ \hdashline
\cite{pavlidou_tsiknakis_2025} &GSR & domain-specific$^\divideontimes$ &LSTM &76.86\\ \hdashline
\cite{ji_zhao_li_2023} &GSR &domain-specific$^\divideontimes$ &RF &80.40\\ \hdashline
\cite{kachele_thiam_amirian_werner_2015} &GSR &domain-specific$^\divideontimes$ &RF &81.90\\ \hdashline
\cite{patil_2024} &GSR &domain-specific$^\divideontimes$ &LR &82.36\\ \hdashline
\cite{pouromran_radhakrishnan_2021} &GSR &domain-specific$^\divideontimes$ &SVM &83.30\\ \hdashline
\cite{patil_patil_2024} &GSR & raw &1D CNN, biLSTM  &83.60\\ \hdashline
\cite{gouverneur_li_2021} &GSR &domain-specific$^\divideontimes$ &MLP &84.22\\ \hdashline
\cite{thiam_bellmann_kestler_2019}  &GSR &raw &1D CNN  &84.57\\ \hdashline
\cite{phan_iyortsuun_2023} &GSR &raw &1D CNN, biLSTM &84.80\\ \hdashline


\multirow{2}{*}{\cite{lu_ozek_kamarthi_2023}} &
\multirow{2}{*}{\begin{tabular}{@{}c@{}}GSR\end{tabular}} &
\multirow{2}{*}{raw} &
1D CNN & \multirow{2}{*}{\begin{tabular}{@{}c@{}}85.56\end{tabular}}\\
& & & Transformer & \\ \hdashline

\multirow{2}{*}{\cite{li_luo_2024}} &
\multirow{2}{*}{\begin{tabular}{@{}c@{}}GSR\end{tabular}} &
\multirow{2}{*}{raw} &
1D CNN & \multirow{2}{*}{\begin{tabular}{@{}c@{}}86.21\end{tabular}}\\
& & & Transformer & \\ \hdashline

\rowcolor{mygray}Our &GSR &raw\textsuperscript{\ding{95}} &Transformer &\textbf{88.99}\\

\bottomrule 
\end{tabular}
\begin{tablenotes}
\scriptsize
\item $\divideontimes$: numerous features \space \ding{72}: pseudo heart rate gain (estimated from videos) \space \ding{70}: PSD \space \ding{96}: waveform-PSD (Concat) \space \ding{95}: waveform \space SVM: Support Vector Machines \space LR: Logistic Regression

\end{tablenotes}
\end{threeparttable}
\end{center}
\end{table}

\renewcommand{\arraystretch}{1.2}
\begin{table}
\caption{Comparison of multimodal-based studies utilizing \textit{BioVid (Part-A)}, NP vs. P\textsubscript{4} task and LOSO cross-validation.}
\label{table:multimodal_based}
\begin{center}
\begin{threeparttable}
\begin{tabular}{ P{0.69cm} P{2.1cm}  P{2.2cm} P{1.2cm} P{0.6cm} }
\toprule
\multirow{2}[2]{*}{\shortstack{Study}}
&\multirow{2}[2]{*}{\shortstack{Modality}}
&\multicolumn{2}{c}{Method} 
&\multirow{2}[2]{*}{\shortstack{Acc\%}}\\
\cmidrule(lr){3-4}
&&Features &ML &\\
\midrule
\midrule
\cite{martinez_picard_2018_b}  &ECG, GSR &domain-specific$^\divideontimes$ &SVM &72.20\\ \hdashline
\cite{werner_hamadi_2014}  &ECG,  EMG, GSR &domain-specific$^{\divideontimes}$ &RF &74.10\\ \hdashline
\cite{pavlidou_tsiknakis_2025} &ECG, EMG, GSR & domain-specific$^\divideontimes$ &LSTM &77.21\\ \hdashline


\multirow{3}{*}{\cite{werner_hamadi_2014}} &
\multirow{3}{*}{\begin{tabular}[c]{@{}c@{}}Video$^{1}$, ECG$^{2}$,\\ EMG$^{2}$, GSR$^{2}$\end{tabular}} &
facial landmarks$^{1}$ & \multirow{3}{*}{RF} & \multirow{3}{*}{\begin{tabular}{@{}c@{}}77.80\end{tabular}}\\
& & 3D distances$^{1}$ & & \\
& & domain-specific$^{2\divideontimes}$ & & \\ \hdashline


\multirow{3}{*}{\cite{kachele_werner_2015}} &
\multirow{3}{*}{\begin{tabular}[c]{@{}c@{}}Video$^{1}$, ECG$^{2}$,\\ GSR$^{2}$\end{tabular}} &
facial landmarks$^{1}$ & \multirow{3}{*}{RF} & \multirow{3}{*}{\begin{tabular}{@{}c@{}}78.90\end{tabular}}\\
& & 3D distances$^{1}$ & & \\
& & domain-specific$^{2\divideontimes}$ & & \\ \hdashline

\cite{gkikas_tachos_2024} &Video$^1$, ECG$^2$ &raw$^1$, heart rate$^2$ &Transformer &82.74\\ \hdashline


\multirow{3}{*}{\cite{kachele_thiam_amirian_werner_2015}} &
\multirow{3}{*}{\begin{tabular}[c]{@{}l@{}}Video$^{1}$, ECG$^{2}$,\\ EMG$^{2}$, GSR$^{2}$\end{tabular}} &
geometric$^{1}$ & \multirow{3}{*}{RF} & \multirow{3}{*}{\begin{tabular}{@{}c@{}}83.10\end{tabular}}\\
& & appearance$^{1}$ & & \\
& & domain-specific$^{2}$ & & \\ \hdashline

\cite{patil_2024} &ECG, EMG, GSR &domain-specific &LR &83.20\\ \hdashline
\cite{wang_xu_2020} &ECG, EMG, GSR &domain-specific &biLSTM &83.30\\ \hdashline
\cite{thiam_kestler_schenker_2020_b} &ECG, EMG, GSR&raw &DDCAE &83.99\\ \hdashline


\multirow{2}{*}{\cite{thiam_hihn_braun_2021}} &
\multirow{2}{*}{\begin{tabular}{@{}c@{}}ECG, EMG, GSR\end{tabular}} &
\multirow{2}{*}{raw} &
DDCAE, & \multirow{2}{*}{\begin{tabular}{@{}c@{}}84.25\end{tabular}}\\
& & & NN & \\ \hdashline

\cite{thiam_bellmann_kestler_2019}  &ECG, EMG, GSR &raw &2D CNN &84.40\\ \hdashline
\cite{jiang_rosio_2024} &GSR, ECG &domain-specific$^{\divideontimes}$ &NN &84.58\\ \hdashline


\multirow{2}{*}{\cite{patil_patil_2024}} &
\multirow{2}{*}{\begin{tabular}{@{}c@{}}Video, GSR\end{tabular}} &
\multirow{2}{*}{raw} &
2D CNN & \multirow{2}{*}{\begin{tabular}{@{}c@{}}84.80\end{tabular}}\\
& & & biLSTM & \\ \hdashline

\multirow{2}{*}{\cite{phan_iyortsuun_2023}} &
\multirow{2}{*}{\begin{tabular}{@{}c@{}}ECG, GSR\end{tabular}} &
\multirow{2}{*}{raw} &
1D CNN & \multirow{2}{*}{\begin{tabular}{@{}c@{}}84.80\end{tabular}}\\
& & & biLSTM & \\ \hdashline

\cite{kachele_thiam_amirian_2016} &ECG, EMG, GSR &domain-specific &RF &85.70\\ \hdashline
\cite{bellmann_Schwenker_2020} &ECG, EMG, GSR &domain-specific &RF &85.80\\ \hdashline


\multirow{2}{*}{\cite{zhi_yu_2019}} &
\multirow{2}{*}{\begin{tabular}{@{}c@{}}Video$^{1}$, ECG$^{2}$,\\ EMG$^{2}$, GSR$^{2}$\end{tabular}} &
facial descriptors$^{1}$ & \multirow{2}{*}{RF} & \multirow{2}{*}{\begin{tabular}{@{}c@{}}86.00\end{tabular}}\\
& & domain-specific$^{2}$ & & \\ \hdashline

\cite{jiang_li_he_2024} &GSR, ECG &domain-specific$^{\divideontimes}$ &NN &87.06\\
\hdashline
\rowcolor{mygray}Our &Video\textsuperscript{\ding{66}}, GSR\textsuperscript{\ding{93}} &raw &Transformer &\textbf{89.08}\\
\bottomrule 
\end{tabular}
\begin{tablenotes}
\scriptsize
\item \ding{66}: RGB-thermal-depth \space \ding{93}: waveform \space $\divideontimes$: numerous features \space DDCAE: Deep Denoising Convolutional Autoencoders \space NN: Neural Network

\end{tablenotes}
\end{threeparttable}
\end{center}
\end{table}

%
%
%

\renewcommand{\arraystretch}{1.2}
\begin{table}
\caption{Comparison of studies on the testing set of \textit{AI4Pain} dataset.}
\label{table:ai4pain_test}
\begin{center}
\begin{threeparttable}
\begin{tabular}{ P{0.99cm} P{0.4cm} P{0.45cm} P{0.3cm} P{0.3cm} P{0.3cm} P{0.6cm} P{1.1cm} P{0.6cm}}
\toprule
\multirow{2}[2]{*}{\shortstack{Study}}
&\multicolumn{6}{c}{Modality}
&\multirow{2}[2]{*}{\shortstack{ML}}
&\multirow{2}[2]{*}{\shortstack{Acc\%}}\\
\cmidrule(lr){2-7}
&Video &fNIRS &EDA &BVP &Resp &SpO$_2$\\
\midrule
\midrule
\cite{gkikas_tsiknakis_painvit_2024}$^\dagger$ &\checkmark &\checkmark &-- &-- &-- &-- &Transformer     &46.67\\ \hdashline
\cite{prajod_schiller_2024}$^\dagger$          &\checkmark &--         &-- &-- &-- &-- &2D CNN          &49.00\\ \hdashline

\multirow{2}{*}{\cite{vianto_2025}$^\dagger$}                
&\multirow{2}{*}{\checkmark} &\multirow{2}{*}{\checkmark} &\multirow{2}{*}{--} &\multirow{2}{*}{--} &\multirow{2}{*}{--} &\multirow{2}{*}{--} 
&CNN-Transformer &\multirow{2}{*}{51.33}\\ \hdashline

\cite{khan_aziz_2025}$^\dagger$                &--         &\checkmark &-- &-- &-- &-- &ENS             &53.66\\ \hdashline
\cite{nguyen_yang_2024}$^\dagger$              &\checkmark &--         &-- &-- &-- &-- &Transformer     &55.00\\  \midrule

\cite{gkikas_kyprakis_resp_2025}$^\ddagger$ &-- &-- &-- &-- &\checkmark &-- &Transformer &42.24 \\ \hdashline

\cite{gkikas_kyprakis_multimodal_2025}$^\ddagger$ &-- &-- &\checkmark &\checkmark &\checkmark &\checkmark &MoE     &54.89\\ \hdashline

\cite{gkikas_kyprakis_eda_2025}$^\ddagger$ &-- &-- &\checkmark &-- &-- &-- &Transformer  &55.17\\\midrule

\rowcolor{mygray}
& -- & \checkmark & -- & -- & -- & -- &  & 52.60\\
\rowcolor{mygray}
& \checkmark & -- & -- & -- & -- & -- &  & 53.67\\
\rowcolor{mygray}
\multirow{-3}{*}{\cellcolor{mygray}Our$^\dagger$}
& \checkmark & \checkmark & -- & -- & -- & -- &
\multirow{-3}{*}{\cellcolor{mygray}Transformer} & \textbf{55.69}\\

\bottomrule 
\end{tabular}
\begin{tablenotes}
\scriptsize
\item BVP: Blood Volume Pulse \space Resp: Respiration Signal \space SpO$_2$: Peripheral Oxygen Saturation  ENS: Ensemble Classifier \space MoE: Mixture of Experts \space $\dagger$: AI4PAIN-First Multimodal Sensing Grand Challenge $\ddagger$: AI4PAIN-Second Multimodal Sensing Grand Challenge

\end{tablenotes}
\end{threeparttable}
\end{center}
\end{table}

\section{Interpretation} 
\label{interpretation}
\begin{figure}
\begin{center}
\includegraphics[scale=0.25]{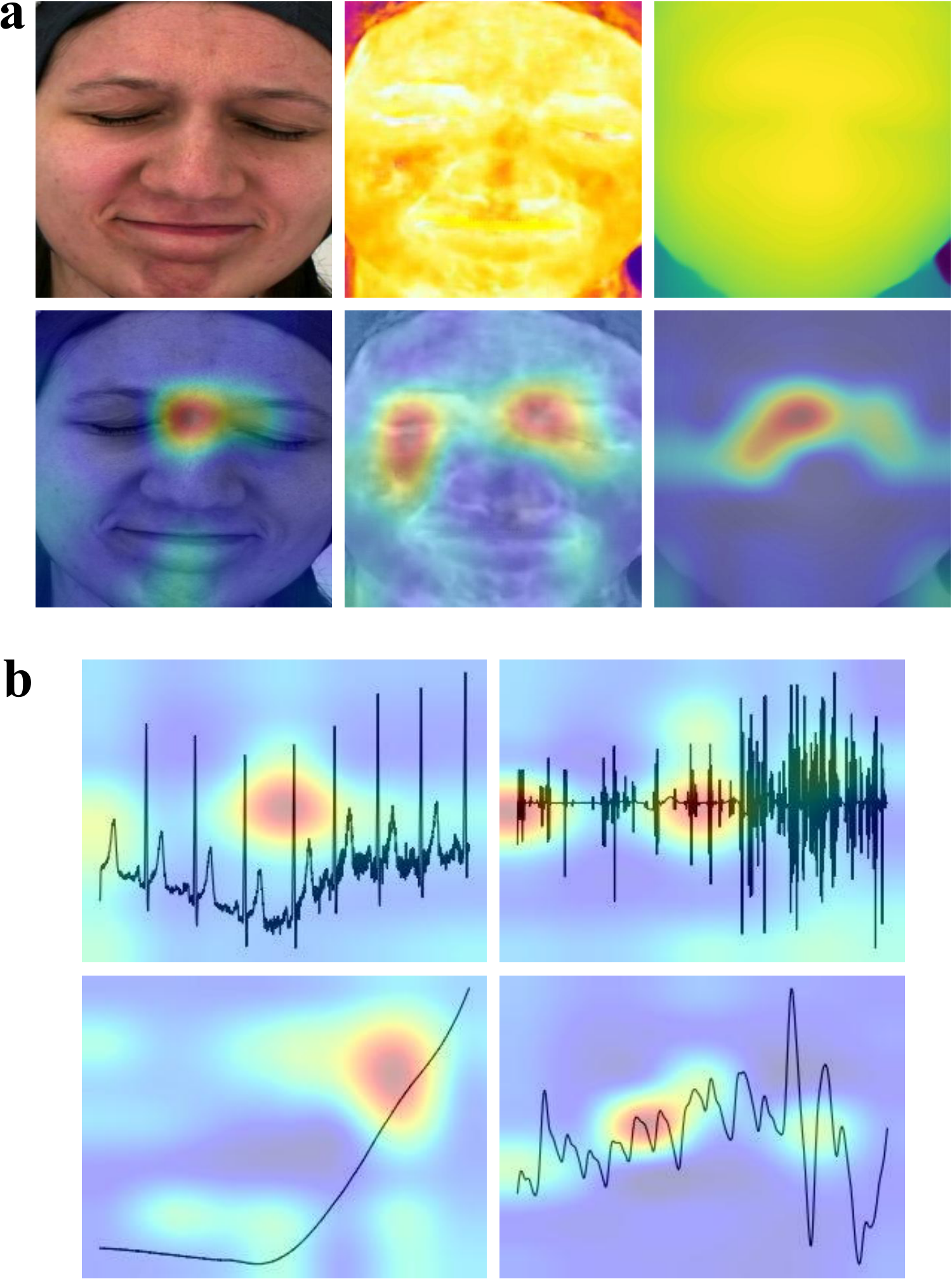} 
\end{center}
\caption{
Attention maps from the \textit{PainFormer}: \textbf{(a)(1\textsuperscript{st} row)} frames from RGB, thermal, and depth video modalities; \textbf{(a)(2\textsuperscript{nd} row)} corresponding attention maps; \textbf{(b)(1\textsuperscript{st} row)} attention maps for ECG and EMG; \textbf{(b)(2\textsuperscript{nd} row)} attention maps for EDA and fNIRS modalities.}
\label{attention_maps}
\end{figure}

Enhancing the interpretability of models is essential for their acceptance and effective integration into clinical settings. In this study, the \textit{PainFormer} has been used to generate attention maps, as shown in Fig. \ref{attention_maps}.
The weights from the \textquotedblleft Stage 4\textquotedblright \space self-attention heads have been applied by interpolating them onto the input images, allowing us to visualize the model's attention areas.

In \hyperref[attention_maps]{Fig. 5(a)}, (1\textsuperscript{st} row), we showcase examples from the RGB, thermal, and depth modalities, and in \hyperref[attention_maps]{Fig. 5(a)}, (2\textsuperscript{nd} row), we present the corresponding attention maps. 
Observations indicate that the model primarily focuses on the glabella region (the area between the eyebrows) in the RGB frame, a zone for manifesting facial expressions. Additional focus is seen on the mental protuberance area (the chin), which is also associated with expressions of pain.
For the thermal frame, the model concentrates on areas around the eyes and the left and right parts of the mouth. Interestingly, these areas correspond to brighter colors in the thermal imagery, indicating higher temperatures rather than direct facial expressions. This suggests that temperature variations influence the model's attention in the thermal frame rather than facial movements and expressions.
In the depth frame, the model targets areas showcasing variations in depth, particularly across the horizontal eye region. There is also slight attention to the frame's lower left and right edges, illustrating depth differences in body parts other than the face, indicating a nuanced understanding of the model's representation of depth.

The ECG attention maps shown in \hyperref[attention_maps]{Fig. 5(b)}, (top left), emphasize primarily a distinct R peak in the trace's center. Notable attention is also directed towards the T waves, especially in the first part of the signal and the T waves that follow the central R peak.
Interestingly, in the EMG attention maps of \hyperref[attention_maps]{Fig. 5(b)}, (top right), the \textit{PainFormer} primarily focuses on the initial and middle sections of the signals. Despite the presence of a muscle contraction burst later in the sequence, the model exhibits less attention to this portion. 
The \textit{Silent-EMG} dataset \cite{gaddy_klein_2020}, on which the \textit{PainFormer} was pre-trained, could relate to this observation. This pre-training background might influence the model's attention and responsiveness to specific sections of the EMG signals.
For the GSR signal in  \hyperref[attention_maps]{Fig. 5(b)}, (bottom left), mild attention is observed at the onset of the response, marking the beginning of the conductance increase. The most intense attention is near the peak amplitude, where the conductance reaches its  maximum level.
For the fNIRS signal in \hyperref[attention_maps]{Fig. 5(b)} (bottom right), the attention map predominantly highlights regions where attention colors align with peaks and rapid changes in HbO2 levels. Notable attention is concentrated in the left, middle, and right sections of the map, where distinct peaks and dips in the signal are observed. This indicates that the \textit{PainFormer} consistently focuses on significant fluctuations in the HbO2 signal, which are likely correlated with pain conditions. Additionally, regions with lower or moderate attention correspond to parts of the time series with stable or minor variations in HbO2, reflecting lower levels of brain activation typically associated with mild or no pain responses.

\section{Computational Resources and Runtime} 
The development of \textit{Painformer} and all presented experiments was conducted on a workstation running Ubuntu 22.04 LTS, with Python 3.9.18, Torch 1.11.0, and CUDA 12.9.
The hardware configuration featured an NVIDIA RTX 3090 GPU paired with an AMD Ryzen 9 5950X 16-core CPU.
Pre-training the foundation model using the proposed multi-task learning approach took approximately $6$ months, while pain-related tasks required $2$-$7$ days, depending on the number of modalities and the selected fusion strategy.

\section{Discussion} 
\label{discussion}
In this study, we introduced \textit{PainFormer}, a vision foundation model designed for pain assessment tasks across various input modalities. \textit{PainFormer} leverages a vision-transformer architecture, pre-trained on $14$ tasks/datasets encompassing $10.9$ million samples using a multi-task learning approach.
The foundation model is supported by supplementary models: \textit{Embedding-Mixer}, a transformer-based model that processes extracted embeddings for pain assessment, and \textit{Video-Encoder}, which encodes video embeddings into a lower-dimensional space. This specific setup facilitates the effective integration of behavioral and physiological modalities.
Furthermore, our approach was evaluated using two pain-related datasets, \textit{BioVid} and \textit{AI4Pain}. We explored a wide array of representations, including RGB, synthetic thermal, depth videos, and waveforms and spectrograms for biosignals such as ECG, EMG, GSR, and fNIRS. Utilizing this diverse set of modalities, we developed multiple pipelines in both unimodal and multimodal settings to assess the quality and effectiveness of the embeddings created by \textit{PainFormer}.

The experiments revealed that the RGB video modality outperformed other video modalities, such as thermal and depth, achieving $76.29\%$ accuracy in the \textit{BioVid} dataset. Notably, thermal and depth also yielded robust results, with accuracies of $71.55\%$ and $71.67\%$, respectively. When combining these video representations, we noted subtle yet important improvements in performance compared to using RGB alone.
Interestingly, the fusion of thermal and depth modalities enhanced accuracy by $1.35\%$, reaching $73.02\%$, which is closely aligned with RGB's performance, suggesting a possible equivalence.
It is crucial to emphasize the significance of this finding, as synthetic thermal and depth estimation videos serve as an intermediary between the highly informative and effective RGB modality, which poses privacy concerns, and facial descriptions such as facial action units, which, while offering high privacy due to the absence of visible faces, tend to yield mediocre performance \cite{werner_hamadi_2014, werner_2016}.
Regarding the physiological modalities, the ECG demonstrated solid performance across all tested representations, with the \textit{spectrogram-PSD} achieving the highest accuracy at $75.49\%$. However, combining various ECG representations did not yield any improvement.
The notably challenging EMG modality performed exceptionally well, achieving over $72\%$ accuracy with \textit{waveform} and \textit{spectrogram-PSD} representations. Interestingly, combining these representations resulted in improvements exceeding $0.5\%$.
GSR signals, recognized as the most effective modality for pain assessment, achieved the highest results compared to other modalities, reaching an $88.9\%$ accuracy with \textit{waveform} representations.
Crucially, our approach to integrating behavioral and physiological modalities with the \textit{BioVid} dataset resulted in substantial performance improvements. By combining GSR embeddings with video embeddings from RGB, thermal, and depth modalities, we achieved an $89.08\%$ accuracy. 
The experiments conducted with the \textit{AI4Pain} dataset showed strong performances across the available modalities. Specifically, using RGB videos, an accuracy of $53.67\%$ was achieved, higher than the $52.60\%$ recorded with fNIRS, while the combination of modalities yielded an increase to $55.69\%$, demonstrating that data fusion, with proper optimization, can indeed enhance performance.

Creating attention maps to interpret how \textit{PainFormer} processes inputs to generate embeddings yielded intriguing results. The foundation model consistently focuses on regions of interest. For instance, RGB videos emphasize distinct facial expressions, while thermal videos target areas displaying more vivid colors that correspond to higher temperatures.
Regarding the depth videos, the analysis shows that the model focuses primarily on the eye area, which is prominent in terms of depth variations. Similarly, it highlights other body areas that exhibit notable depth differences compared to the head.
The situation is analogous to the biosignals, where \textit{PainFormer} can detect subtle yet significant variations across all inputs, including ECG, EMG, GSR, and fNIRS. This capability suggests that the model can capture a comprehensive representation of pain-related data, beneficial for research and clinical applications. However, it is important to note that further investigation is necessary. The pre-training process might bias the model's focus towards specific areas of interest, potentially leading to conflicts and inconsistencies with actual pain events.

Finally, the results are intriguing when comparing the proposed approach to existing methods in the literature. Across datasets and various modalities, our performances are consistently state-of-the-art. Our method ranks among the best reported using the \textit{BioVid} dataset, particularly for video-based approaches. However, it falls short of methods integrating facial features with cardiac-related information extracted directly from the videos \cite{yang_guan_yu_2021, huang_dong_2022}.
Interestingly, our method excels with biosignals, including ECG, EMG, and GSR, achieving the highest recorded performances.
\textit{PainFormer} also consistently delivers high results in unimodal and multimodal settings, underscoring its robustness regardless of the input and the pipelines.
When using the \textit{AI4Pain} dataset, our multimodal approach achieved the highest results reported to date. 
However, it should be noted that this dataset is relatively new, so further studies and extensive research are needed.

\section{Conclusion} 
\label{conclusion}
This study presented \textit{PainFormer}, a vision transformer-based foundation model for pain recognition tasks. Pre-trained on $10.9$ million samples across various modalities and datasets, including facial recognition, emotion recognition, and biosignal datasets. This makes it the first general-purpose and foundation model in automatic pain assessment documented in the literature.
Extensive evaluations across various input modalities and representations, including RGB, synthetic thermal, depth videos, and biosignals like ECG, EMG, GSR, and fNIRS, have demonstrated that the proposed model consistently extracts high-quality feature representations, regardless of the input. A direct comparison with $75$ methodologies from the literature using two datasets exhibited that the model delivers state-of-the-art performance across all modalities in unimodal and multimodal configurations.
Moreover, by generating attention maps, we offered insights into the foundation model's functioning, highlighting the specific focus areas within the inputs.
Finally, we recommend that future research explore multi-modality approaches, which have proven to be the most effective for assessing pain phenomena in real-world settings. Additionally, synthetic data shows considerable promise, and further exploration is needed. Developing methods for interpretation is also essential, especially for the potential integration of these frameworks into clinical practice.

\bibliographystyle{IEEEtran}
\bibliography{library}

\begin{IEEEbiography}
[{\includegraphics[width=1in,height=1.25in,clip,keepaspectratio]{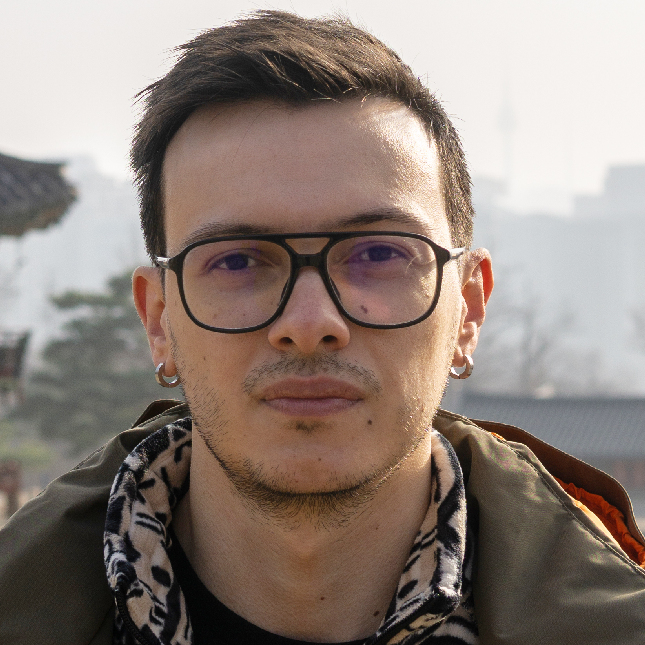}}]
{Stefanos Gkikas}
received his M.Sc. in Computer Vision from the University of Burgundy, France, in 2020.
In 2025, he earned his Ph.D. from the School of Electrical and Computer Engineering at the Hellenic Mediterranean University.
His doctoral research focused on pain recognition, with the dissertation titled \textit{\textquotedblleft A Pain Assessment Framework Based on Multimodal Data and Deep Machine Learning Methods\textquotedblright}.
Throughout his Ph.D., he specialized in the analysis of videos and biosignals.
He is now a postdoctoral researcher at the Computational Biomedicine Laboratory of FORTH/ICS.
His research interests include affective computing, health monitoring systems, and computational physiology.
\end{IEEEbiography}

\begin{IEEEbiography}
[{\includegraphics[width=1in,height=1.25in,clip,keepaspectratio]{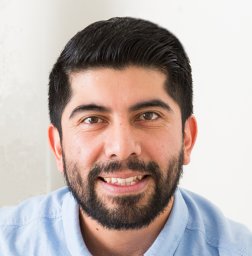}}]
{Raul Fernandez Rojas}
received his Ph.D. from the University of Canberra, Australia, in 2018, specialising in biomarker identification through computational methods and cognitive engineering. From 2018 to 2019, he served as a Postdoctoral Research Associate at the University of New South Wales, within the Australian Defence Force Academy. He is now an Associate Professor in the Faculty of Science and Technology at the University of Canberra. His research interests encompass signal processing, embedded systems design, affective computing, and multimodal sensing.
\end{IEEEbiography}
\begin{IEEEbiography}
[{\includegraphics[width=1in,height=1.25in,clip,keepaspectratio]{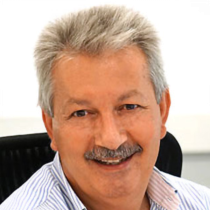}}]
{Manolis Tsiknakis}
received his M.Sc (1983) and Ph.D. (1989) degrees from the University of Bradford, UK. He performed postdoctoral training at the University of Bradford (1990-1991) and the Institute of Computer Science at FORTH (1992).
Since 2012, he has been a Professor of Biomedical Informatics and eHealth at the Department of Electrical and Computer Engineering at the Hellenic Mediterranean University and a visiting Professor at the Computational Biomedicine Laboratory of FORTH/ICS. His main areas of expertise include approaches for semantic health data integration and interoperability of health information systems, affective computing and its application in developing smart eHealth solutions, and service platforms for pervasive eHealth and mHealth services.
\end{IEEEbiography}

\end{document}